\documentclass{article}

\usepackage{arxiv}
\usepackage[utf8]{inputenc} 
\usepackage[T1]{fontenc}    
\usepackage[colorlinks=true, citecolor=blue,urlcolor=blue]{hyperref}       
\usepackage{url}            
\usepackage{booktabs}       
\usepackage{amsfonts}       
\usepackage{nicefrac}       
\usepackage{microtype}      
\usepackage{graphicx}
\usepackage{fontawesome}
\usepackage[title]{appendix}
\usepackage[font=footnotesize,labelfont=bf]{caption}   
\newcommand{\veclatin}[1]{\bm{#1}} 

\newcommand{\vecgreek}[1]{\pmb{#1}}

\usepackage{amsmath, amsthm, amssymb}
\usepackage{dsfont}
\usepackage{mathtools}
\usepackage{calc}
\usepackage[flushleft]{threeparttable}
\DeclareMathOperator*{\argmax}{argmax}
\usepackage{bm} 
\usepackage{algorithm}
\usepackage{algpseudocode}
\usepackage{etoolbox}
\usepackage{tikz}
\usetikzlibrary{tikzmark}
\usetikzlibrary{calc}

\errorcontextlines\maxdimen

\newcommand{\ALGtikzmarkcolor}{black}
\newcommand{\ALGtikzmarkextraindent}{4pt}
\newcommand{\ALGtikzmarkverticaloffsetstart}{-.5ex}
\newcommand{\ALGtikzmarkverticaloffsetend}{-.5ex}
\makeatletter
\newcounter{ALG@tikzmark@tempcnta}

\newcommand\ALG@tikzmark@start{%
	\global\let\ALG@tikzmark@last\ALG@tikzmark@starttext%
	\expandafter\edef\csname ALG@tikzmark@\theALG@nested\endcsname{\theALG@tikzmark@tempcnta}%
	\tikzmark{ALG@tikzmark@start@\csname ALG@tikzmark@\theALG@nested\endcsname}%
	\addtocounter{ALG@tikzmark@tempcnta}{1}%
}

\def\ALG@tikzmark@starttext{start}
\newcommand\ALG@tikzmark@end{%
	\ifx\ALG@tikzmark@last\ALG@tikzmark@starttext
	\else
	\tikzmark{ALG@tikzmark@end@\csname ALG@tikzmark@\theALG@nested\endcsname}%
	\tikz[overlay,remember picture] \draw[\ALGtikzmarkcolor] let \p{S}=($(pic cs:ALG@tikzmark@start@\csname ALG@tikzmark@\theALG@nested\endcsname)+(\ALGtikzmarkextraindent,\ALGtikzmarkverticaloffsetstart)$), \p{E}=($(pic cs:ALG@tikzmark@end@\csname ALG@tikzmark@\theALG@nested\endcsname)+(\ALGtikzmarkextraindent,\ALGtikzmarkverticaloffsetend)$) in (\x{S},\y{S})--(\x{S},\y{E});%
	\fi
	\gdef\ALG@tikzmark@last{end}%
}

\apptocmd{\ALG@beginblock}{\ALG@tikzmark@start}{}{\errmessage{failed to patch}}
\pretocmd{\ALG@endblock}{\ALG@tikzmark@end}{}{\errmessage{failed to patch}}
\makeatother

\usepackage{setspace}
\let\Algorithm\algorithm
\renewcommand\algorithm[1][]{\Algorithm[#1]\setstretch{1.4}}

\algdef{SE}[SUBALG]{Indent}{EndIndent}{}{\algorithmicend\ }%
\algtext*{Indent}
\algtext*{EndIndent}

\usepackage[authoryear, round]{natbib}
\title{\textbf{CatBoostLSS} \\ \vspace{0.5em} {\large An extension of CatBoost to probabilistic forecasting}} 			

\author{Alexander März$^{*}$\thanks{$^{*}$Address for correspondence: \texttt{alex.maerz@gmx.net.}
}
  	\vspace{-1em}
}

\begin{document} 
	
\maketitle

\vspace{2em} 

\begin{abstract}
We propose a new framework of CatBoost that predicts the entire conditional distribution of a univariate response variable. In particular, \texttt{CatBoostLSS} models all moments of a parametric distribution (i.e., mean, location, scale and shape [LSS]) instead of the conditional mean only. Choosing from a wide range of continuous, discrete and mixed discrete-continuous distributions, modelling and predicting the entire conditional distribution greatly enhances the flexibility of CatBoost, as it allows to gain insight into the data generating process, as well as to create probabilistic forecasts from which prediction intervals and quantiles of interest can be derived. We present both a simulation study and real-world examples that demonstrate the benefits of our approach. 
\end{abstract}


\vspace{2em}

\section{Introduction} \label{sec:introduction}

To reason rigorously under uncertainty we need to invoke the language of probability \citep{Zhang.2020}. Any model that falls short of providing quantification of the uncertainty attached to its outcome is likely to yield an incomplete and potentially misleading picture. While this is an irrevocable consensus in statistics, a common misconception, albeit a very persistent one, is that machine learning models usually lack proper ways of quantifying uncertainty. Despite the fact that the two terms exist in parallel and are used interchangeably, the perception that machine learning and statistics imply a non-overlapping set of techniques remains lively, both among practitioners and academics. This is vividly portrayed by the provocatively (and potentially tongue-in-cheek) statement of Brian D. Ripley that '\textit{machine learning is statistics minus any checking of models and assumptions}' that he made during the useR! 2004, Vienna conference that served to illustrate the difference between machine learning and statistics \citep{Zeileis.2016}. 

In fact, the relationship between statistics and machine learning is artificially complicated by such statements and is unfortunate at best, as it implies a profound and qualitative distinction between the two disciplines \citep{Januschowski.2020}. The paper by \citet{Breiman.2001} is a noticeable exception, as it proposes to differentiate the two based on scientific culture, rather than on methods alone. Both statistics and machine learning create models from data, but for different purposes. There is the statistical culture that is well embedded in statistical theory and that assumes the data to be generated by a stochastic process. The aim is to draw inference from the sample and to provide insights into the data generating process. Hence, while the emphasize of statistics is on inference, mathematical rigour and elegance, model validity and estimation of model parameters, the algorithmic culture of machine learning is rather concerned with out-of-sample fit, computational performance and function optimization \citep{Januschowski.2020}. 

While the approaches discussed in \citet{Breiman.2001} are an admissible partitioning of the space of how to analyse and model data, more recent advances have gradually made this distinction less clear-cut (see Section \ref{sec:research} and the references therein for an overview). In fact, the current research trend in both statistics and machine learning gravitates towards bringing both disciplines closer together. In an era of increasing necessity that the output of prediction models needs to be turned into explainable and reliable insights, this is an exceedingly promising and encouraging development, as both disciplines have much to learn from each other. Along with \citet{Januschowski.2020}, we argue that it is more constructive to seek common ground than it is to introduce artificial boundaries. As such, this paper contributes to further closing the gap between the two cultures by extending statistical boosting to a machine learning approach that accounts for for all distributional properties of the data. In particular, we present an extension of CatBoost, which has gained much popularity and attention recently as a competitor to the eminent XGBoost model. We term our model \texttt{CatBoostLSS}, as it combines the accuracy and speed of CatBoost with the flexibility and interpretability of GAMLSS that allow for the estimation and prediction of the entire conditional distribution $F_{Y}(y|\mathbf{x})$.\footnote{We follow \citet{Hothorn.2018} and denote $\mathbb{P}(Y \leq y | \mathbf{X} = \mathbf{x}) = F_{Y}(y|\mathbf{x})$ the conditional distribution of a potentially continuous, discrete or mixed discrete-continuous response $Y$ given explanatory variables $\mathbf{X} = \mathbf{x}$.} \texttt{CatBoostLSS} allows the user to choose from a wide range of continuous, discrete and mixed discrete-continuous distributions to better adapt to the data at hand, as well as to provide predictive distributions, from which prediction intervals and quantiles can be derived. \texttt{CatBoostLSS} therefore contributes to the growing literature on statistical machine learning that aims at weakening the separation between the 'Data Modelling Culture' and 'Algorithmic Modelling Culture', so that models designed mainly for prediction can also be used to describe and explain the underlying data generating process of the response of interest.\footnote{In its essence, statistical machine learning formulates basic concepts of machine learning from a statistical perspective, where the link between inference and computational efficiency is central. More generally, statistical machine learning is concerned with the development of algorithms that learn from observed data by assuming a stochastic data model which can then be used for inference, predictions and for making decisions \citep{Hutter.2008}.} 

The remainder of this paper is organised as follows: Section \ref{sec:gamlss} introduces the reader to distributional modelling and Section \ref{sec:research} presents an overview of related research. In Section \ref{sec:catboostlss}, we formally introduce \texttt{CatBoostLSS}, while Section \ref{sec:applications} presents both a simulation study and real world examples that provide a walk-through of the functionality of our model. Section \ref{sec:implementation} gives an overview of available software implementations and Section \ref{sec:conclusion} concludes.

\section{Distributional Modelling} \label{sec:gamlss}

\begin{quote} 
	\it{The ultimate goal of regression analysis is to obtain information about the \textbf{[entire] conditional distribution} of a response given a set of explanatory variables}.\footnote{Emphasize added.}\citep{Hothorn.2014}
\end{quote}

Consulting the literature on machine learning shows that the main focus so far has been on prediction accuracy and estimation speed. In fact, even though machine learning approaches (e.g., Random Forest or Gradient Boosting-type algorithms) outperform many statistical models when it comes to prediction accuracy, the output/forecast of these models provides information about the conditional mean $\mathbb{E}(Y|\mathbf{X} = \mathbf{x})$ only. As a consequence, this class of models is rather reluctant to reveal other characteristics of the (predicted) distribution and falls short in applications where probabilistic forecasts are required, e.g., for assessing prediction uncertainty in form of prediction intervals. By focusing on point-forecasts and hoping them to materialize, we ignore one of the fundamental principles of nature which is that, by default, future is uncertain and the best we can hope for when providing forecasts is to properly quantify the uncertainty that is attached to them. In contrast, probabilistic forecasts are predictions in the form of a probability distribution, rather than simply a single point estimate. Having estimated and predicted the conditional distribution, we can create sample paths that can be interpreted as a possible realization of the future. In this context, the introduction of Generalised Additive Models for Location Scale and Shape (GAMLSS) by \citet{Rigby.2005} has stimulated a lot of research and culminated in a new branch of statistics that focuses on modelling the entire conditional distribution as functions of covariates. This section introduces the reader to the general idea of distributional modelling.\footnote{In order to fully comprehend their beauty and elegance, we draw the reader's attention to \citet{Rigby.2005, Klein.2015b, Klein.2015c, Stasinopoulos.2017}.}

In its original formulation, GAMLSS assume that a univariate response follows a distribution $\mathcal{D}$ that depends on up to four parameters, i.e., $y_{i} \stackrel{ind}{\sim} \mathcal{D}(\mu_{i}, \sigma^{2}_{i}, \nu_{i}, \tau_{i}), i=1,\ldots,n$, where $\mu_{i}$ and $\sigma^{2}_{i}$ are location and scale parameters, respectively, while $\nu_{i}$ and $\tau_{i}$ correspond to shape parameters such as skewness and kurtosis. Hence, the framework allows to model not only the mean (or location) but all parameters as functions of explanatory variables. In contrast to Generalised Linear (GLM) and Generalised Additive Models (GAM), the assumption of the response belonging to an exponential family type of distribution is relaxed in GAMLSS and replaced by a general distribution family, including highly skewed and/or kurtotic continuous,  discrete and mixed discrete distributions. While the original formulation of GAMLSS in \citet{Rigby.2005} suggests that any distribution can be described by location, scale and shape parameters, it is not necessarily true that the distribution at hand is actually characterized by parameters that represent shape parameters, i.e., skewness and kurtosis. Hence, we follow \citet{Klein.2015b} and use the term distributional modelling and GAMLSS interchangeably. From a frequentist point of view, distributional modelling can be formulated as follows

\begin{equation}
y_{i} \stackrel{ind}{\sim} \mathcal{D}\Big(h_{1}(\theta_{i1}) = \eta_{i1}, h_{2}(\theta_{i2}) = \eta_{i2}, \ldots, h_{K}(\theta_{iK}) = \eta_{iK}\Big), \qquad i = 1, \ldots, n,
\end{equation} 

\noindent where $\mathcal{D}$ denotes a parametric distribution for the response $\textbf{y} = (y_{1}, \ldots, y_{n})^{\prime}$ that depends on $K$ distributional parameters $\theta_{k}, k = 1, \ldots, K$, and with $h_{k}(\cdot)$ denoting a known monotonic function relating the distribution parameters to a predictor $\vecgreek{\eta}_{k}$. In its most generic form, the predictor $\vecgreek{\eta}_{k}$ is given by

\begin{equation}
\vecgreek{\eta}_{k} = f_{k}(\veclatin{x}), \qquad k = 1, \ldots, K \label{eq:gamlss}
\end{equation} 

Within the original distributional regression framework, the functions $f_{k}(\cdot)$ usually represent an additive GAM-type predictor that are based on a basis function approach using splines, i.e., $\vecgreek{\eta}_{k} = \textbf{x}_{k}\vecgreek{\beta}_{k} + \sum^{p_{k}}_{j=1}f_{k,j}(\veclatin{z}_{j})$, where $\vecgreek{\beta}_{k} = (\beta_{1k}, \beta_{2k}, \ldots, \beta_{qk})^{\prime}$ is a parameter vector modelling linear effects or categorical variables, $\textbf{x}_{k}$ is the corresponding design matrix and $f_{k,j}(\veclatin{z}_{j})$ reflect different types of regression effects that model the effect of a continuous covariate $\veclatin{z}_{j}$.\footnote{See \citealt{Fahrmeir.2011} and \citealt{Fahrmeir.2013} for further details.} However, it is important to stress that besides its classic representation, the predictor specification in Equation \eqref{eq:gamlss} is generic enough to also represent classification and regression trees, which allows us to extend CatBoost to a probabilistic framework. Concerning the estimation of distributional regression, it relies on the availability of first and second order derivatives of the (log)-likelihood function needed for Fisher-scoring type algorithms. As we will see in Section \ref{sec:catboostlss}, this is very closely related to the estimation of CatBoost, which we will exploit to arrive at \texttt{CatBoostLSS}.

We would like to draw the attention of the reader to an implication that is a consequence of modelling and predicting the entire distribution and that has received relatively little interest in the machine learning community until very recently \citep{Quinonero.2009}. Many machine learning algorithms have been proposed and shown to be very successful. One assumption required to guarantee performance for prediction tasks is the test data to have the same distribution as the training data, or more formally, that train and test observations to be independent and identically distributed (iid) realizations arising from the same stationary distribution $y \stackrel{iid}{\sim} \mathcal{D}(\veclatin{\theta})$, where $\veclatin{\theta}$ is a vector of distributional parameters. As such, we aim to train a model that, given new inputs of the test set, can accurately predict the corresponding unseen output. In real world applications, however, distributions are complex and likely to be non-stationary, rendering the conditional response distribution $F_{Y}(y|\mathbf{x})$ to remain unchanged very difficult.\footnote{It is important to stress that we focus on a change of the conditional distribution in our discussion only and do not investigate the consequences of a covariate shift on model performance. Covariate shift is a simpler case of the more general dataset shift, where the distribution of the input variables changes (e.g., different geographies, schools, or hospitals may be drawn from different demographics), while the conditional distribution of the outputs given the inputs remains unchanged. For a more detailed exposition on dataset shift in machine learning, we refer the interested reader to \citet{Quinonero.2009}.} If not addressed properly, a shift in the conditional distribution between training and test data may lead to inaccuracy of parameter estimates and instability of predictions \citep{Quinonero.2009}. To illustrate the implications of distributional modelling, let us re-visit the concept of stationarity used in time series analysis, with covariates $\veclatin{x}$ including time. Most forecasting methods assume that the time series at hand can be rendered approximately stationary using appropriate transformations, e.g., difference-stationary or trend-stationary. In general, one can distinguish two forms of stationarity. The first, and the weaker one, is covariance stationarity which requires the first moment (i.e., the mean) and auto-covariance to not vary with respect to time. The second, and stricter one, is strong stationarity that can be formulated as follows

\begin{equation}
F_{Y}(y_{t_{1}}, \ldots, y_{t_{n}}) = F_{Y}(y_{t_{1} + \tau }, \ldots, y_{t_{n} + \tau}), \qquad \forall \mbox{ } n, t_{1}, \ldots, t_{n}, \tau 
\end{equation}

\noindent where $F_{Y}(\cdot)$ is the joint cumulative distribution function of $\left\lbrace y_{t}\right\rbrace $ at times $t$. Given that $F_{Y}(\cdot)$ does not change with a shift in time of $\tau$, it follows that all parameters of a strictly stationary process are time invariant. However, this is a very restrictive assumption that is likely to be violated in many real-world applications. Along with \citet{Quinonero.2009}, we argue that flexible modelling frameworks are essential for the development of a detailed understanding of the problems attached to modelling non-stationary distributions. As we will see in subsequent sections, distributional modelling in general and \texttt{CatBoostLSS} in particular, allows to uncover and analyse the underlying mechanisms of a distribution shift, such as change in variance, by relating all distributional parameters to explanatory variables. In particular, distributional modelling implies that the observations are independent, but not necessarily identical realizations $y \stackrel{ind}{\sim} \mathcal{D}\big(\veclatin{\theta}(\veclatin{x})\big)$, where all distributional parameters $\veclatin{\theta}(\veclatin{x})$ are related to and allowed to change with covariates. \footnote{As all distributional parameters are functions of covariates, distributional modelling is able to account for the non-stationarity so that stationarity does not need to serve as default assumption in applied modelling. For a discussion on non-stationarity modelling in hydrologic flood frequency analyses and climate change modelling see \citet{Villarini.2009}, \citet{Milly.2015} and \citet{Serinaldi.2015}.}

\section{CatBoostLSS} \label{sec:catboostlss} 

In this section, we introduce \texttt{CatBoostLSS}. It is based on the CatBoost (for categorical boosting) algorithm recently introduced by  \citet{Prokhorenkova.2019} and \citet{Dorogush.2018}. There are several characteristics that set CatBoost apart from other existing boosting approaches, namely the implementation of ordered boosting, a permutation-driven alternative to existing approaches, and an efficient algorithm for vector representation of categorical data that makes CatBoost particularly suitable for handling data sets with a lot of categorical features. Both novelties are using random permutations of the training examples to fight the prediction shift caused by a special kind of target leakage present in all existing implementations of gradient boosting algorithms \citep{Prokhorenkova.2019,Dorogush.2018}.\footnote{We refer to \citet{Prokhorenkova.2019, Dorogush.2018} for a more detailed exposition of the functioning of CatBoost.} Even though it provides a flexible interface for parameter tuning, CatBoost outperforms many of the existing state-of-the-art implementations of gradient boosted decision trees, such as XGBoost of \citet{Chen.2016}, on a diverse set of popular tasks, without any parameter tuning using default parameters only.\footnote{See \href{https://catboost.ai/\#benchmark}{https://catboost.ai/benchmark} for a comparison.} It has both a CPU and GPU implementation which are faster than other gradient boosting libraries.\footnote{So far, \texttt{CatBoostLSS} is available for CPU training only. Also, due to CatBoost implementation restrictions, training time of \texttt{CatBoostLSS} is of magnitude 2 to 3 slower compared to the training time of CatBoost.} Depending on the task at hand, CatBoost allows the user to select between gradient and Newton boosting.\footnote{In a recent paper, \citet{Sigrist.2019} provides empirical evidence that Newton Boosting generally outperforms gradient boosting on the majority of data sets used for the comparison. \citet{Sigrist.2019} mainly attributes the advantage of Newton over gradient boosting to the variability in $h_{i}$, i.e., the more variation there is in the second order terms, the more pronounced is the difference between the two approaches and the more likely is Newton to outperform gradient boosting. Also note that if $h_{i}$ is 1 everywhere, Newton and gradient boosting are equivalent. This is the case for, e.g., the squared error loss (hence assuming a Normal distribution with constant variance), i.e., $l[y_{i}, \hat{y}^{(t)}_{i}] = \frac{1}{2}(\hat{y}^{(t)}_{i} - y_{i})^{2}$, we get $g_{i} = (\hat{y}^{(t)}_{i} - y_{i} )$ and $h_{i}$ = 1. As a consequence, if we use any loss function other than squared error loss, Newton tree boosting should outperform gradient boosting.} To establish the connection between GAMLSS and \texttt{CatBoostLSS}, we need to recall that for Newton boosting, the first and second order partial derivatives of the (element-wise) loss function with respect to the fitted label is calculated at each iteration $t$. As such, Newton boosting amounts to a weighted least-squares regression problem at each iteration, which is solved using base learners (e.g., using CART). As a consequence, Newton boosting can be understood as an iterative empirical risk minimization procedure in function space, that determines both the step direction and step length at the same time. This is where \texttt{CatBoostLSS} makes the connection to GAMLSS, as empirical risk minimization and Maximum Likelihood estimation are closely related. Recall from Section \ref{sec:gamlss} that GAMLSS are estimated using the first and second order partial derivatives of the log-likelihood function with respect to the distributional parameter $\theta_{k}$ of interest. By selecting an appropriate loss, or equivalently, a log-likelihood function, Maximum Likelihood can be formulated as empirical risk minimization so that the resulting CatBoost model can be interpreted as a statistical model.\footnote{Note that maximizing the negative log-likelihood is equivalent to minimizing an empirical risk function.} 

Now that we have outlined that \texttt{CatBoostLSS} can be interpreted as a statistical model by having established the connection between the estimation of GAMLSS and CatBoost, we can introduce \texttt{CatBoostLSS} more formally. Algorithm \eqref{alg:catboostlss} gives a conceptual overview of the steps involved to estimate our model. We have designed \texttt{CatBoostLSS} in such a way that the initial CatBoost implementation remains unchanged, so that its full functionality is still available. In a sense, \texttt{CatBoostLSS} is a wrapper around CatBoost, where we interpret the loss function from a statistical perspective by formulating empirical risk minimization as Maximum Likelihood estimation. As outlined in Algorithm \eqref{alg:catboostlss}, we first need to specify an appropriate log-likelihood, from which Gradients and Hessians are derived, that represent the partial first and second order derivatives of the log-likelihood with respect to the distributional parameter $\theta_{k}$ of interest. In contrast, however, to the approach in \citet{Mayr.2012} and \citet{Thomas.2018}, that uses a component-wise gradient descent algorithm, where each of the $\theta_{k}$ is updated successively in each iteration, using the current estimates of the other distribution parameters $\theta_{-k}$ as input, our approach is a two-step procedure. In the first step, we estimate a separate model for each distributional parameter $\theta_{k}, k = 1, \ldots K$, where the unconditional Maximum Likelihood estimates of the parameters $\theta_{-k}$, not currently being estimated, are used as offset values. As such, while $\theta_{k}$ is estimated, $\theta_{-k}$ are treated being constant. Once all $\theta_{k}$ are estimated, we update each parameter by incorporating information from all other parameters until a stopping criterion based on the global deviance is met. Once all parameters are updated and the global deviance has converged, we can draw random samples from the predicted distribution that allows us to create probabilistic forecasts from which prediction intervals and quantiles of interest can be derived. 

\begin{algorithm}[h!]
	\caption{CatBoostLSS} \label{alg:catboostlss}
	\hspace*{\algorithmicindent} \textbf{Input:} Data set $D$ \\ 
	\hspace*{\algorithmicindent} \textbf{Required:} Appropriate (log)-likelihood/loss function $\ell[\cdot]$ \\
	\hspace*{\algorithmicindent} \textbf{Ensure:} Negative Gradient and negative Hessian exist and are non-zero
	\begin{algorithmic}[1]%
		\State \textbf{Step 1:} Estimate distributional parameter $\theta_{i,k}$ independently of other parameters $\theta_{i,-k}$.
		\Indent
		\For{$k$-th distributional parameter $\theta_{k}, k = 1, \ldots, K$}	%
		\State Initialize $\hat{\theta}_{-k} = \argmax_\theta \ln \ell[\mathbf {y}, \theta_{-k}]$  \Comment{Initialize with unconditional ML-estimate} 		%
		\State Define loss function $\ell[y, \hat{f}_{\theta_{k}}, \hat{g}_{\theta_{k}}, \hat{h}_{\theta_{k}}]$
		\State Define evaluation metric $\xi[y, \hat{f}_{\theta_{k}}]$
		\For{$m = 1, \ldots, M$ boosting iterations}	%
		\State $\hat{g}^{m}_{\theta_{k}} = - \left[ \frac{\partial \ell[y, f(x)]}{\partial f(x)} \right]_{f(x) = \hat{f}^{(m-1)}_{\theta_{k}}(x)}$ \Comment{Negative Gradient of $k$-th parameter}%
		\State $\hat{h}^{m}_{\theta_{k}} = - \left[ \frac{\partial^{2} \ell[y, f(x)]}{\partial f(x)^{2}} \right]_{f(x) = \hat{f}^{(m-1)}_{\theta_{k}}(x)}$ \Comment{Negative Hessian of $k$-th parameter}%
		\State Update estimate: $\hat{f}^{(m)}_{\theta_{k}} = \eta \hat{f}^{(m-1)}_{\theta_{k}}$ \Comment{$\eta$ denotes the learning rate}%
		\EndFor	
		\State Output: $\hat{f}_{\theta_{k}} = \sum^{M}_{m=0}\hat{f}^{(m)}_{\theta_{k}}, \quad k = 1, \ldots, K.$
		\EndFor 
		\EndIndent	
		\State \textbf{Step 2:} Using estimated models of Step 1, update $\hat{\theta}_{k}$ with information from $\hat{\theta}_{-k}$.
		\Indent
		\While{diff $\geq \epsilon$ and $q$ $\leq$ max\_iter} %
		\For{$k$-th distributional parameter $\theta_{k}$, $k = 1, \ldots, K$}	%
		\State Repeat steps 6-10 and update $\theta_{k}$ by incorporating information from all other
		\State parameters $\theta_{-k}$:				
		\State \begin{equation}
		\begin{aligned}
		\left(\hat{\theta}^{(q)}_{1}, \ldots, \hat{\theta}^{(q)}_{K}\right)  \quad  & \xrightarrow[]{\text{Update $\theta_{1}$}} \quad \hat{\theta}^{(q+1)}_{1} & \xrightarrow[]{\text{Output}} \quad \hat{f}^{*(q+1)}_{\theta_{1}}, \nonumber \\
		\left(\hat{\theta}^{(q+1)}_{1}, \hat{\theta}^{(q)}_{2}, \ldots, \hat{\theta}^{(q)}_{K}\right) \quad  & \xrightarrow[]{\text{Update $\theta_{2}$}} \quad \hat{\theta}^{(q+1)}_{2} & \xrightarrow[]{\text{Output}} \quad \hat{f}^{*(q+1)}_{\theta_{2}}, \nonumber \\
		\left(\hat{\theta}^{(q+1)}_{1}, \hat{\theta}^{(q+1)}_{2}, \ldots, \hat{\theta}^{(q)}_{K}\right) \quad  &  \xrightarrow[]{\text{Update $\theta_{3}$}} \quad \hat{\theta}^{(q+1)}_{3}  & \xrightarrow[]{\text{Output}} \quad \hat{f}^{*(q+1)}_{\theta_{3}}, \nonumber \\
		\quad &\makebox[\widthof{${}\xrightarrow{}$}][c]{\vdots} \\
		\left(\hat{\theta}^{(q+1)}_{1}, \hat{\theta}^{(q+1)}_{2}, \ldots, \hat{\theta}^{(q)}_{K}\right) \quad & \xrightarrow[]{\text{Update $\theta_{K}$}} \quad \hat{\theta}^{(q+1)}_{K}  & \xrightarrow[]{\text{Output}} \quad \hat{f}^{*(q+1)}_{\theta_{K}}. \nonumber
		\end{aligned}
		\end{equation}					
		\EndFor	
		\State $\mbox{deviance}_{q} \gets -2\ln \ell[\hat{f}^{*(q+1)}_{\theta} \,;\mathbf {y}]$
		\State $\mbox{diff} \gets |(\mbox{deviance}_{q+1}\ - \mbox{deviance}_{q})| / \mbox{deviance}_{q}$
		\State $q \gets q + 1$
		\EndWhile
		\EndIndent
	\end{algorithmic}
	\hspace*{\algorithmicindent} \textbf{Final Output: } $\hat{f}^{*}_{\theta_{k}}, \quad k = 1, \ldots, K.$ 
\end{algorithm}

\newpage

\section{Related Research} \label{sec:research}

Reviewing the current literature at the intersect between machine learning and statistics shows that there has been an incredibly rich stream of ideas that aim at bringing the two disciplines closer together. As this section cannot give an exhaustive overview of all approaches, we refer the interested reader to \citet{Maerz.2019} and the references therein.

Amongst all of the available implementations, the approach closest to \texttt{CatBoostLSS} is the one introduced in our related paper \citet{Maerz.2019}. In fact, \texttt{CatBoostLSS} and \texttt{XGBoostLSS} differ only by the base learner used to model the conditional distribution. The choice of which approach to use depends, as always, on the purpose and problem at hand. \texttt{CatBoostLSS} makes use of CatBoost's efficient algorithm for representing categorical data which makes it particularly suitable for handling data sets where the majority of the features are categorical. Also, the fact that CatBoost achieves state-of-the-art prediction performance with basically no or very little hyper-parameter tuning, makes it particularly useful for distributions with many parameters $\theta_{k}, k = 1, \ldots, K$, as the careful selection of hyper-parameters for learning higher conditional moments, e.g., kurtosis or skewness, is crucial for the stability and convergence of the algorithm. While existing GAMLSS frameworks and implementations are supposed to perform well for small to medium sized data sets, \texttt{CatBoostLSS} plays off its strengths in situations where the user faces data sets with hundreds of thousands or even millions of observations.

\section{Applications} \label{sec:applications} 

In the following, we present both a simulation study and real world examples that demonstrate the functionality of \texttt{CatBoostLSS}. It is important to note that, since CatBoost yields state-of-the-art prediction results without extensive parameter tuning typically required by other machine learning methods, we estimate all \texttt{CatBoostLSS} models in the following using default hyper-parameter settings.\footnote{It is important to note that all available parameter-tuning approaches implemented in CatBoost (e.g., early stopping, CV, etc.) are also available for \texttt{CatBoostLSS}.}

\subsection{Simulation} \label{sec:simulation} 

We start with a simulated a data set that exhibits heteroskedasticity, where the interest lies in predicting the 5\% and 95\% quantiles.\footnote{For the simulation, we slightly modify the example presented in \citet{Hothorn.2018c}.} The dots in red show points that lie outside the 5\% and 95\% quantiles, which are indicated by the black dashed lines.

\begin{figure}[h!]
	\centering
	\includegraphics[width=0.68\linewidth]{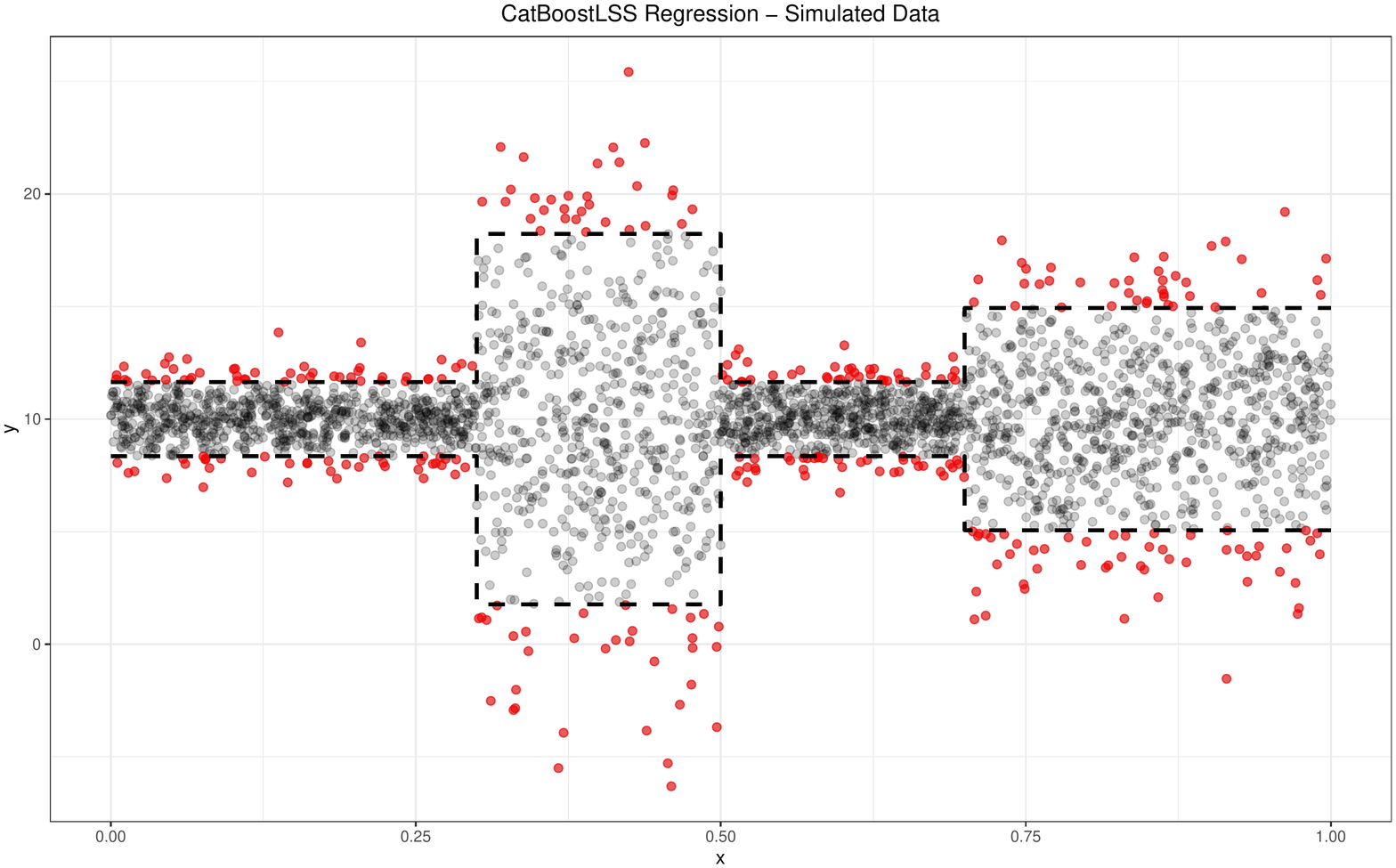}
	\caption{Simulated Train Dataset with 7,000 observations $y \sim \mathcal{N}(10,(1 + 4(0.3 < x < 0.5) + 2(x > 0.7))$. Points outside the 5\% and 95\% quantile are coloured in red. The black dashed lines depict the actual 5\% and 95\% quantiles. Besides the only informative predictor $x$, we have added $X_{1}, \ldots, X_{10}$ as noise variables to the design matrix.}
	\label{fig:sim_data}
\end{figure}

\newpage

\noindent As splitting procedures, that are internally used to construct trees, can detect changes in the mean only, standard implementations of machine learning models are not able to recognize any distributional changes (e.g., change of variance), even if these can be related to covariates \citep{Hothorn.2018c}. As such, CatBoost doesn't provide any uncertainty quantification in its current implementation, as the model focuses on predicting the conditional mean $\mathbb{E}(Y|\mathbf{X} = \mathbf{x})$ only, without any assessment on the full predictive distribution $F_{Y}(y|\mathbf{x})$. This is in contrast to \texttt{CatBoostLSS}, where all distributional parameters are modelled as functions of covariates.  

\begin{figure}[h!]
	\centering
	\includegraphics[width=0.8\linewidth]{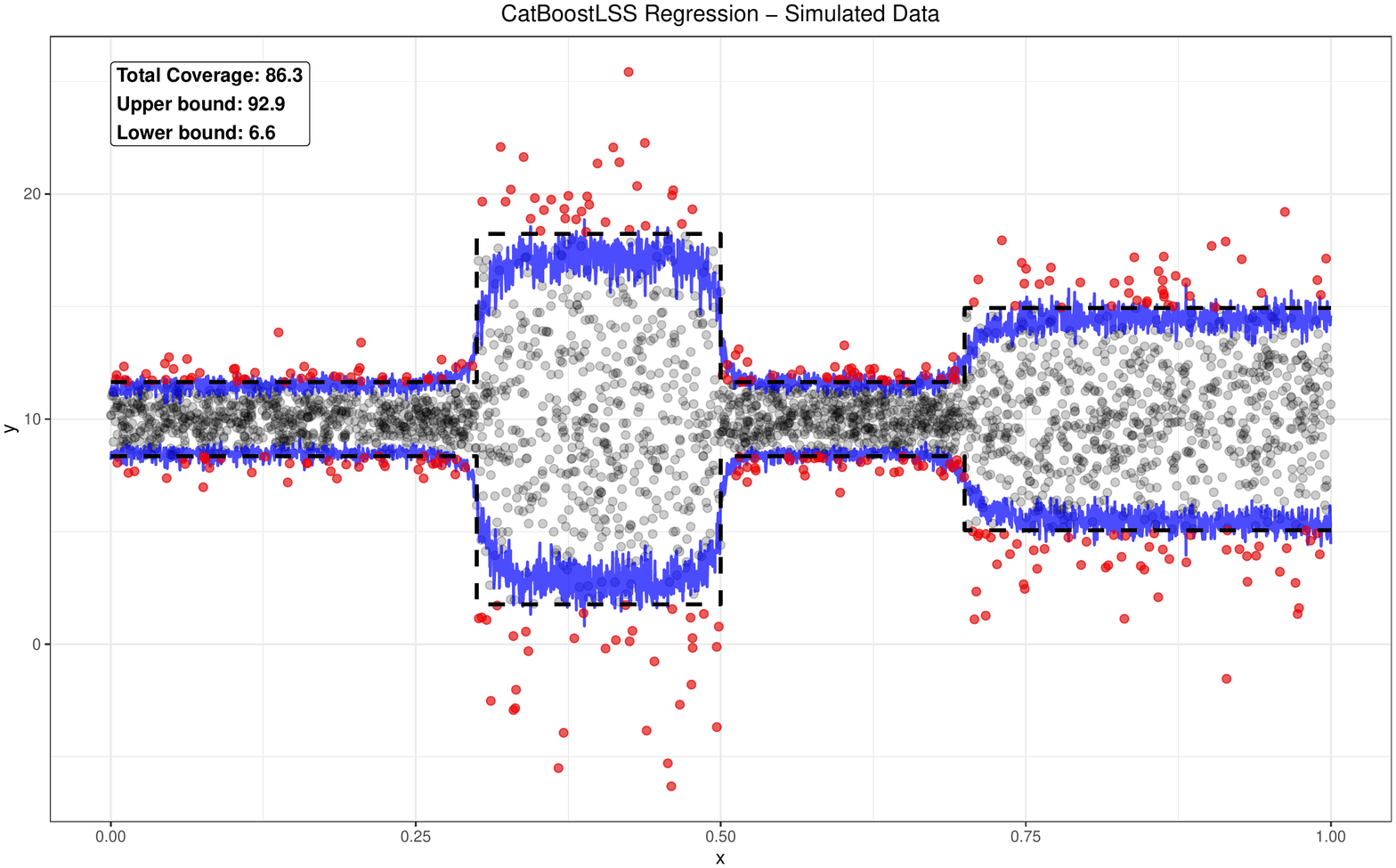}
	\caption{Simulated Test Dataset with 3,000 observations $y \sim \mathcal{N}(10,(1 + 4(0.3 < x < 0.5) + 2(x > 0.7))$. Points outside the conditional 5\% and 95\% quantile are in red. The black dashed lines depict the actual 5\% and 95\% quantiles. Conditional 5\% and 95\% quantile predictions obtained from \texttt{CatBoostLSS} are depicted by the blue lines. Besides the only informative predictor $x$, we have added $X_{1}, \ldots, X_{10}$ as noise variables to the design matrix.}
	\label{fig:sim_mbo}
\end{figure}

Let us now fit \texttt{CatBoostLSS} to the data. In general, the syntax is similar to the original CatBoost implementation. However, the user has to make a distributional assumption by specifying a family in the function call. As the data has been generated by a Normal distribution, we use the Normal as a function input. The user also has the option of providing a list of hyper-parameters that are used for tuning the model. Once the model is trained, we can predict all parameters of the distribution. As \texttt{CatBoostLSS} allows to model the entire conditional distribution, we obtain prediction intervals and quantiles of interest directly from the predicted quantile function. Figure \ref{fig:sim_mbo} shows the predictions of \texttt{CatBoostLSS} for the 5\% and 95\% quantile in blue.

\begin{figure}[h!]
	\centering
	\includegraphics[width=0.7\linewidth]{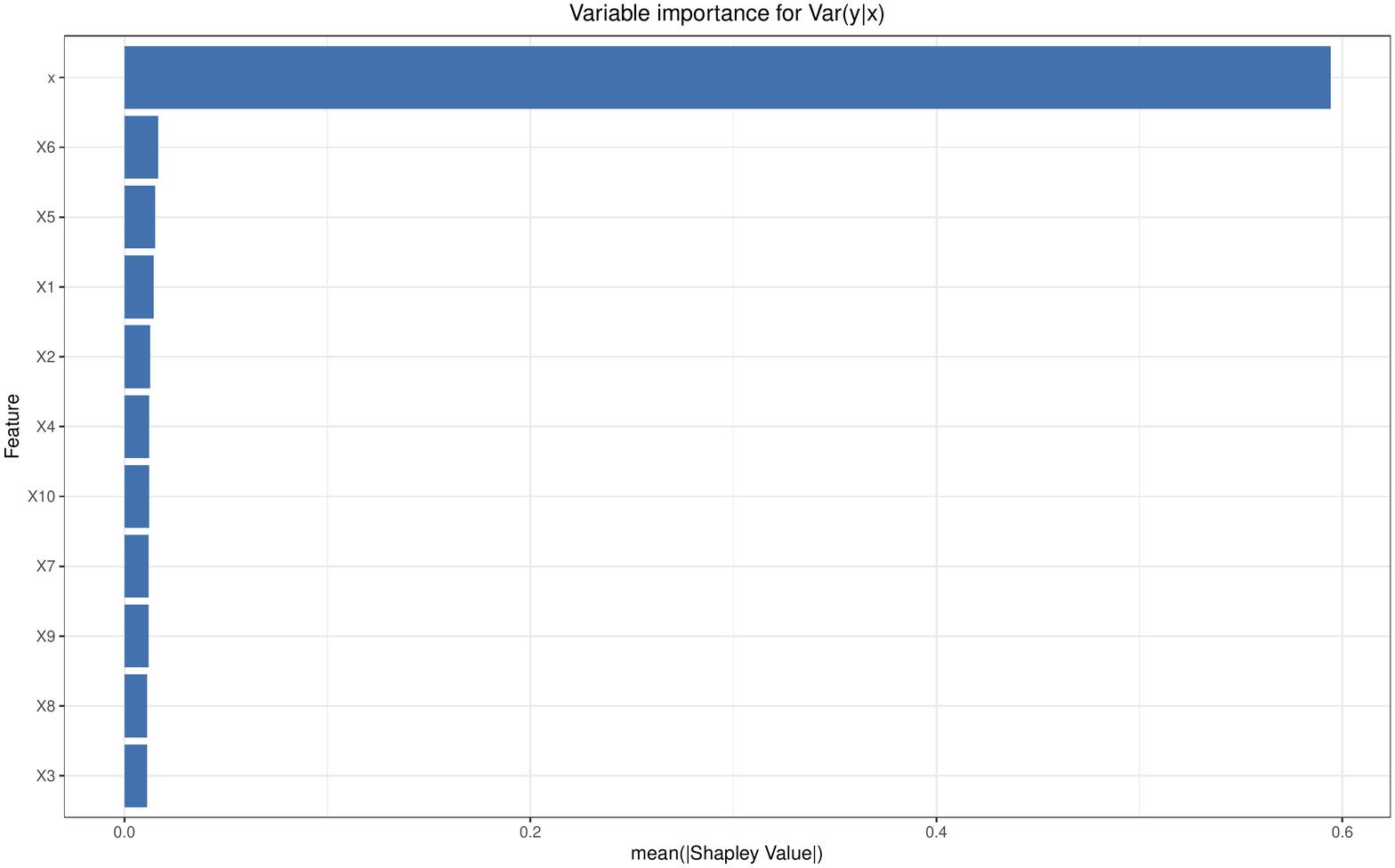}
	\caption{Mean Absolute Shapley Value of $\mathbb{V}(Y|\mathbf{X} = \mathbf{x})$.}
	\label{fig:shap_value}
\end{figure}

\noindent Comparing the coverage of the intervals with the nominal level of 90\% shows that \texttt{CatBoostLSS} does not only correctly model the heteroskedasticity in the data, but it also provides a reasonable forecast for the 5\% and 95\% quantiles. The flexibility of \texttt{CatBoostLSS} also comes from its ability to provide attribute importance, as well as partial dependence plots, for all of the distributional parameters. In the following we only investigate the effect on the conditional variance. Figure \ref{fig:shap_value} shows that \texttt{CatBoostLSS} has identified the only informative predictor $x$ and does not consider any of the noise variables $X_{1}, \ldots, X_{10}$ as important features. 

\newpage

\begin{figure}[h!]
	\centering
	\includegraphics[width=0.7\linewidth]{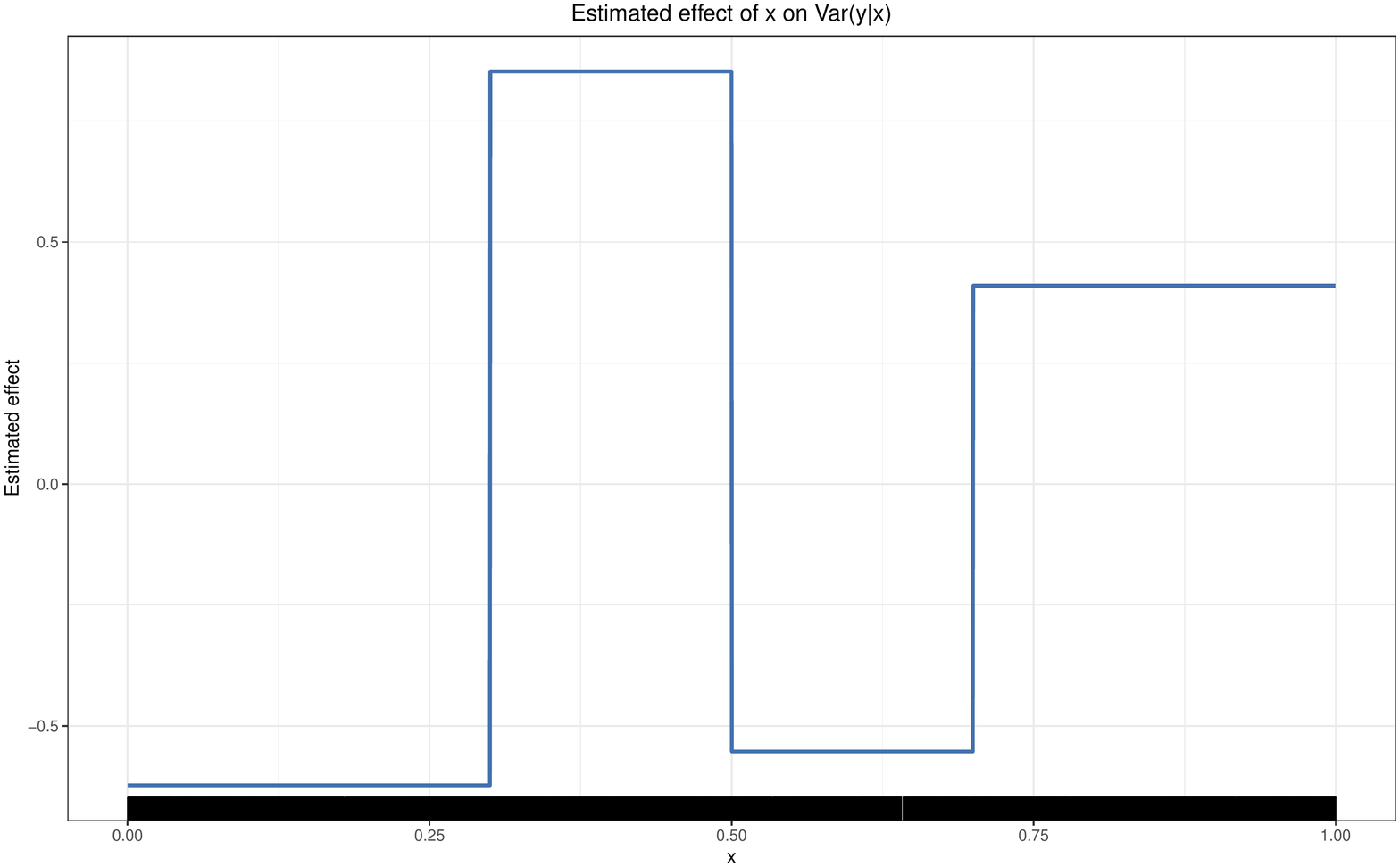}
	\caption{Smoothed Partial Dependence Plot of $\mathbb{V}(Y|\mathbf{X} = \mathbf{x})$.}
	\label{fig:part_effec}
\end{figure}

\noindent Inspecting partial dependence plots of $\mathbb{V}(Y|\mathbf{X} = \mathbf{x})$ shown in Figure \ref{fig:part_effec} indicates that it also correctly identifies the heteroskedasticity in the data.

\subsection{Munich Rent}

\begin{figure}[h!]
	\centering
	\includegraphics[width=0.8\linewidth]{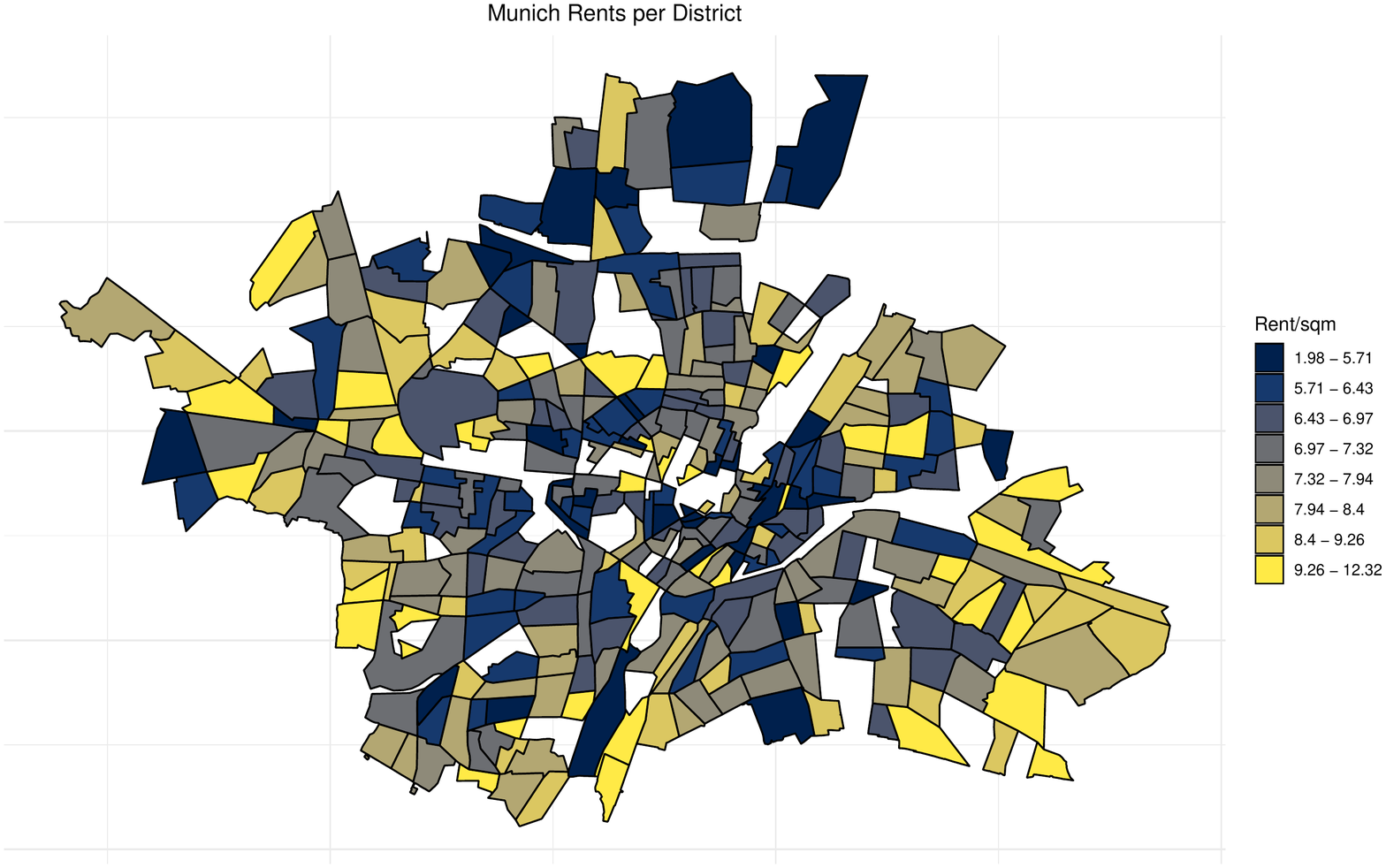}
	\caption{Munich Rents per square meter per district.}
	\label{fig:rent_map}
\end{figure}

Considering there is an active discussion around imposing a freeze in German cities on rents, we have chosen to re-visit the famous Munich Rent data set commonly used in the GAMLSS literature, as Munich is among the most expensive cities in Germany when it comes to living costs. In this example, we illustrate the functionality of \texttt{CatBoostLSS} using a sample of 2,053 apartments from the data collected for the preparation of the Munich rent index 2003, as shown in Figure \ref{fig:rent_map}. As our dependent variable, we select \emph{Net rent per square meter in EUR}.

The first decision one has to make is about choosing an appropriate distribution for the response. As there are many potential candidates, we use an automated approach based on the generalised Akaike information criterion (GAIC). 

\begin{table}[h!]
	\begin{center}
		\caption{Candidate Response Distributions}
		\scalebox{0.9}{%
			\begin{threeparttable}
				
				\begin{tabular*}{0.5\textwidth}{l @{\extracolsep{\fill}}r}
					\toprule
					Distribution  & GAIC    \\
					\midrule
					GB2     & 6588.29 \\
					NO      & 6601.17 \\
					GG      & 6602.02 \\
					BCCG    & 6602.26 \\
					WEI     & 6602.37 \\
					exGAUS  & 6603.17 \\
					BCT     & 6603.35 \\
					BCPEo   & 6604.26 \\
					GA      & 6707.85 \\
					GIG     & 6709.85 \\
					LOGNO   & 6839.56 \\
					IG      & 6871.12 \\
					IGAMMA  & 7046.50 \\
					EXP     & 9018.04 \\
					PARETO2 & 9020.04 \\
					GP      & 9020.05 \\
					\bottomrule
				\end{tabular*}
				\begin{tablenotes}
					\tiny
					\item \noindent Generalized Beta Type 2 (GB2); Normal (NO); Generalized Gamma (GG); Box-Cox Cole and Green (BCCG); Weibull (WEI); ex-Gaussian (exGAUS); Box-Cox t-distribution (BCT); Box-Cox Power Exponential (BCPEo); Gamma (GA); Generalized Inverse Gaussian (GIG); Log-Normal (LOGNO); Inverse Gaussian (IG); Inverse Gamma (IGAMMA); Exponential (EXP); Pareto Type 2 (PARETO2); Generalized Pareto (GP).
				\end{tablenotes}
		\end{threeparttable}}
		\label{tab:dist}
	\end{center}
\end{table}

\noindent Even though Table \ref{tab:dist} suggests the Generalized Beta Type 2 to provide the best approximation to the data, we use the more parsimonious Normal distribution, as it has only two distributional parameters, compared to 4 of the Generalized Beta Type 2. In general, though, \texttt{CatBoostLSS} is flexible to allow the user to choose from a wide range of continuous, discrete and mixed discrete-continuous distributions. The good fit of the Normal distribution is also confirmed by the the density plot, where the response of the train data is presented as a histogram, while the fitted Normal is shown in red. 

\begin{figure}[h!]
	\centering
	\includegraphics[width=0.7\linewidth]{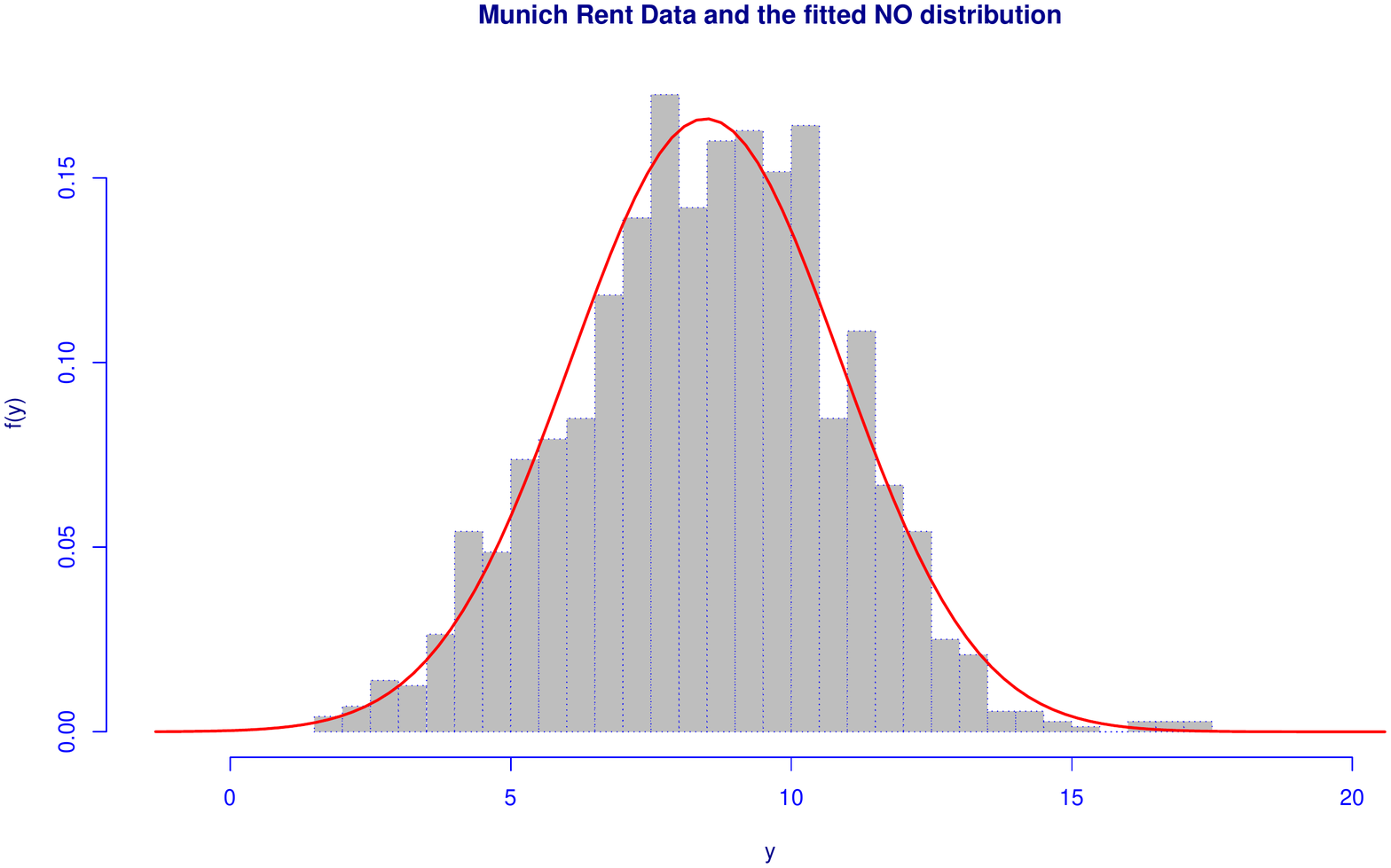}
	\caption{Fitted Normal Distribution.}
	\label{fig:fitdist}
\end{figure}

\begin{figure}[h!]
	\centering
	\includegraphics[width=0.9\linewidth]{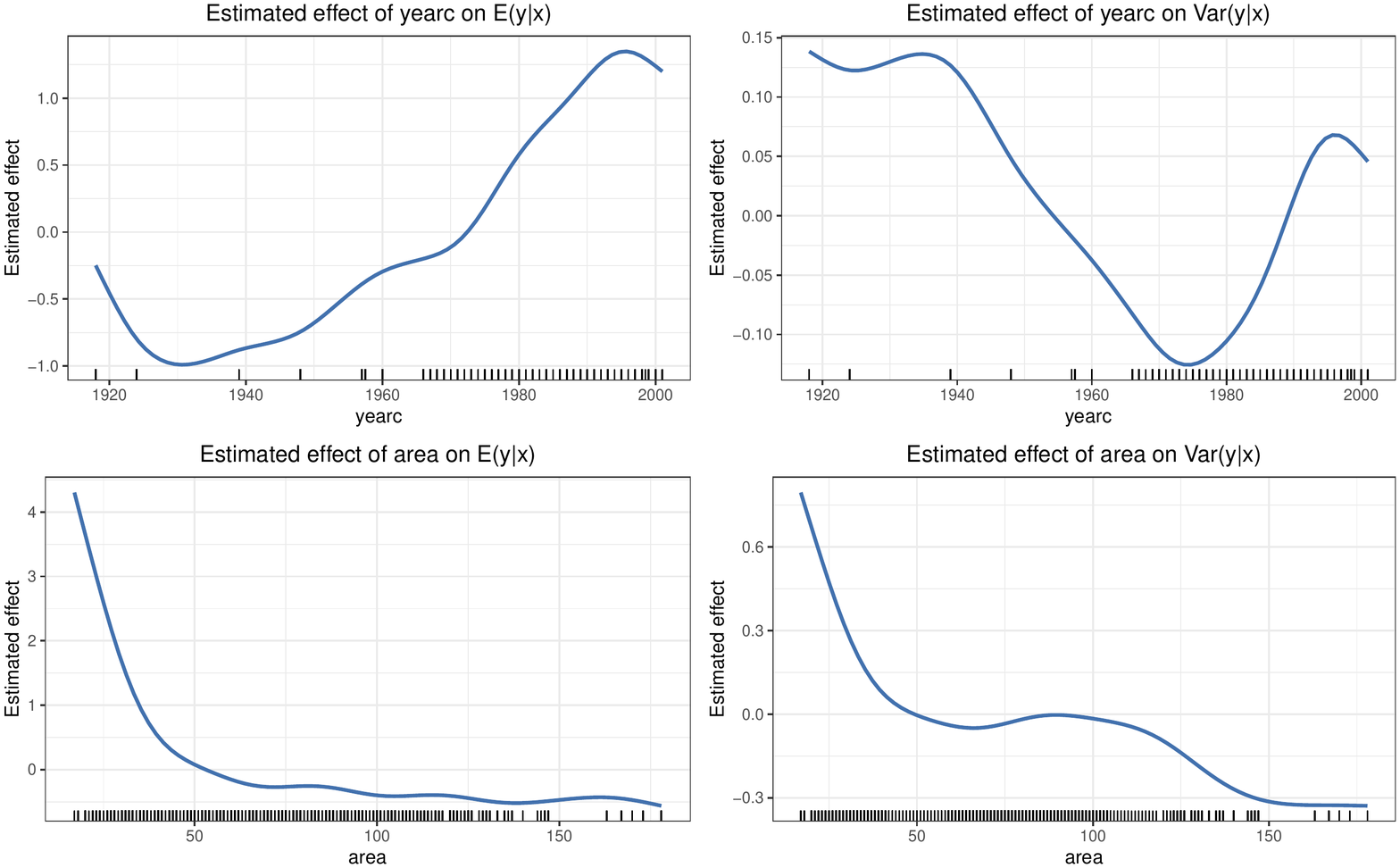}
	\caption{Estimated Partial Effects.}
	\label{fig:pdp}
\end{figure}

\noindent Now that we have specified the distribution, we fit our \texttt{CatBoostLSS} model to the data. Looking at the estimated effects presented in Figure \ref{fig:pdp} indicates that newer flats are on average more expensive, with the variance first decreasing and increasing again for flats built around 1980 and later. Also, as expected, rents per square meter decrease with an increasing size of the apartment.

\newpage

\begin{figure}[h!]
	\centering
	\includegraphics[width=0.6\linewidth]{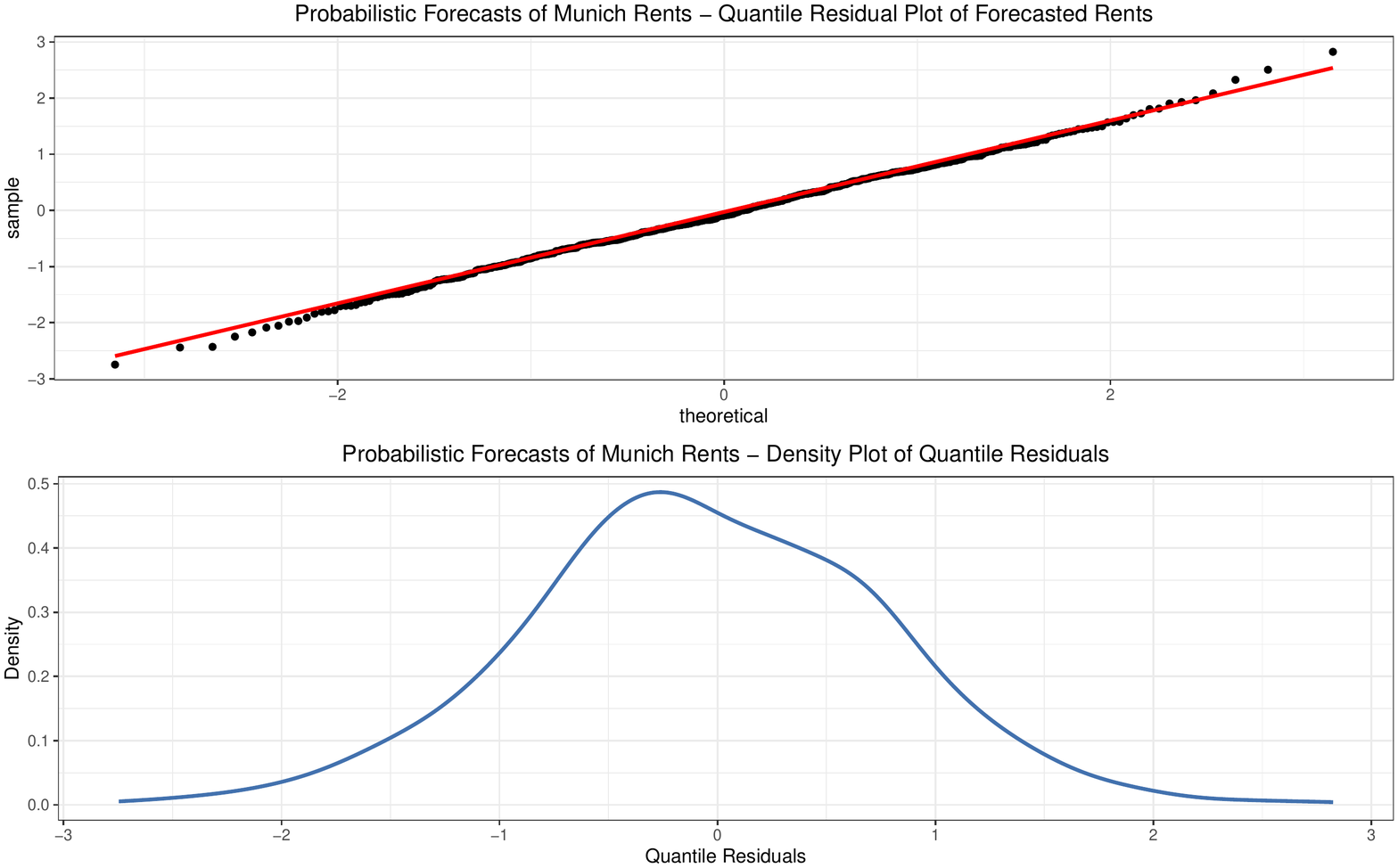}
	\caption{Quantile Residuals.}
	\label{fig:quantres}
\end{figure}

\noindent The diagnostics for \texttt{CatBoostLSS} are based on quantile residuals of the fitted model and are shown in Figure \ref{fig:quantres}.\footnote{For continuous response data, the quantile residuals are based on $u_{i} = F_{i}(y_{i}|\hat{\veclatin{\theta}})$, where $F_{i}(\cdot)$ is the cumulative distribution function estimated for the $i$-th observation, $\hat{\veclatin{\theta}}$ contains all estimated parameters and $y_{i}$ is the corresponding observation. If $F_{i}(\cdot)$ is close to the true distribution of $y_{i}$, then $u_{i}$ approximately follows a uniform distribution. The quantile residuals are then defined as $\hat{r}_{i} = \phi^{-1}(u_{i})$, where $\phi^{-1}(\cdot)$ is the inverse cumulative distribution function of the standard Normal distribution. Hence, ${r}_{i}$ is approximately standard Normal if the estimated model is close to the true one.}

\begin{figure}[h!]
	\centering
	\includegraphics[width=0.7\linewidth]{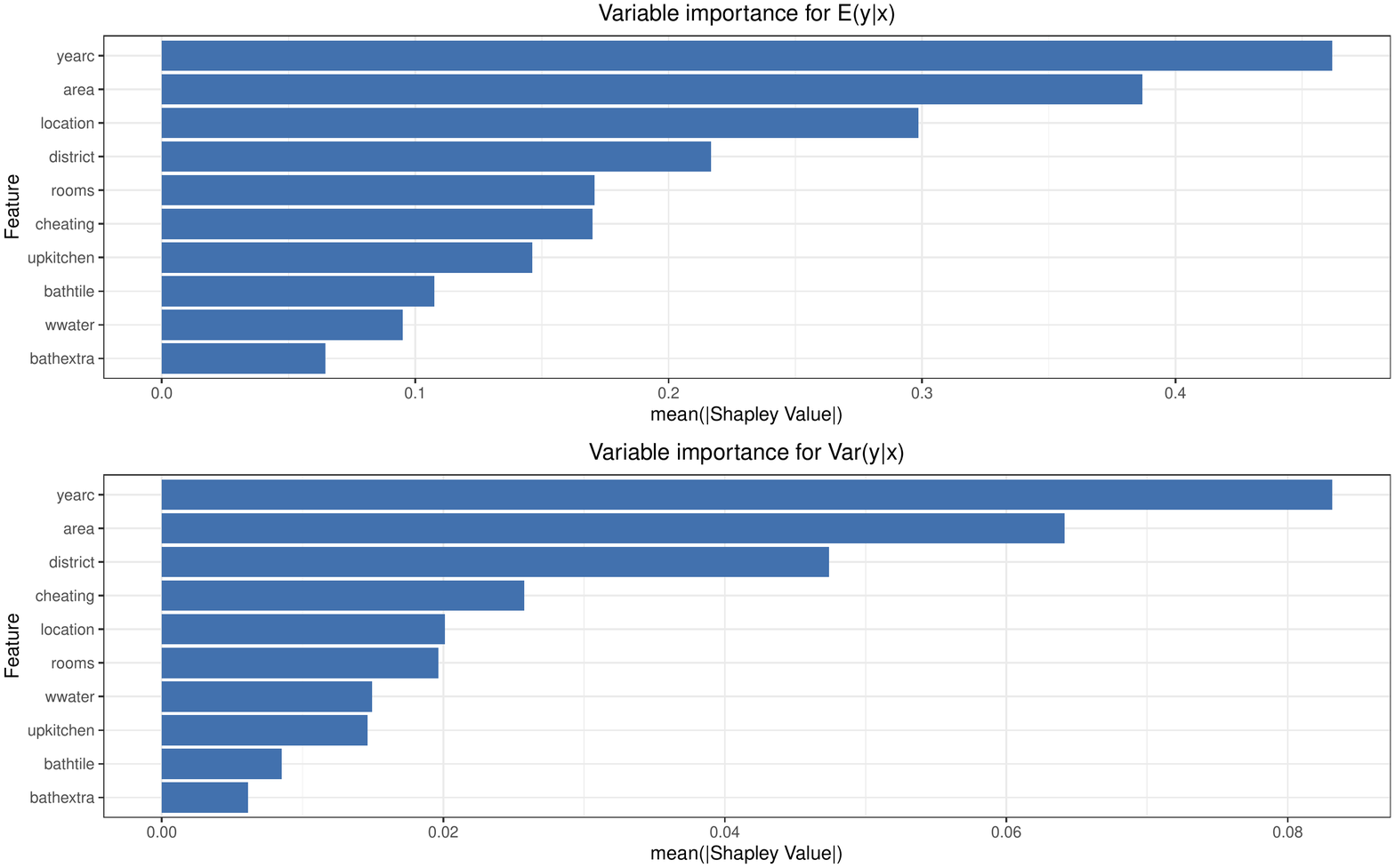}
	\caption{Mean Absolute Shapley Value of  $\mathbb{E}(Y|\mathbf{X} = \mathbf{x})$ and $\mathbb{V}(Y|\mathbf{X} = \mathbf{x})$.}
	\label{fig:munich_shap}
\end{figure}

\noindent \texttt{CatBoostLSS} provides a well calibrated forecast and confirms that our model is a good approximation to the data. \texttt{CatBoostLSS} also allows to investigate feature importance for all distributional parameters. Looking at the top 10 features with the highest Shapley values for both the conditional mean and variance in Figure \ref{fig:munich_shap}  indicates that both $yearc$ and $area$ are considered as being the most important variables.

\newpage

\noindent To get a more detailed overview of which features are most important for our model, we can also plot the SHAP values of every feature for every sample. The plot below sorts features by the sum of SHAP value magnitudes over all samples and uses SHAP values to show the distribution of the impacts each feature has on the model output. The colour represents the feature value (red high, blue low). This reveals for example that newer flats increase rents on average.

\begin{figure}[h!]
	\centering
	\includegraphics[width=0.6\linewidth]{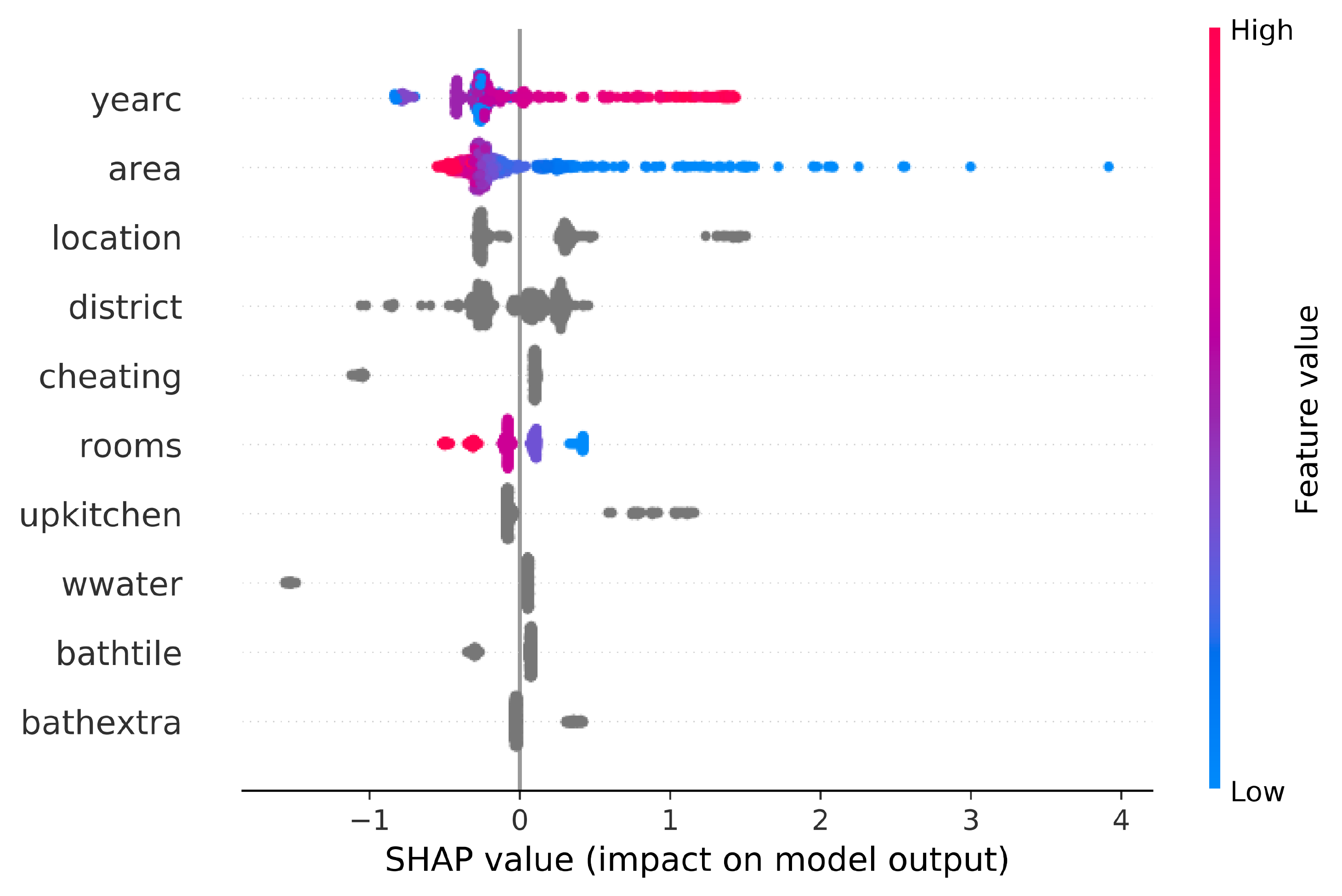}
	\caption{Shapley Values of $\mathbb{E}(Y|\mathbf{X} = \mathbf{x})$.}
	\label{fig:munich_shap_all}
\end{figure}

\noindent We can also visualize all predictions and assess the attribute importance.

\begin{figure}[h!]
	\centering
	\includegraphics[width=0.7\linewidth]{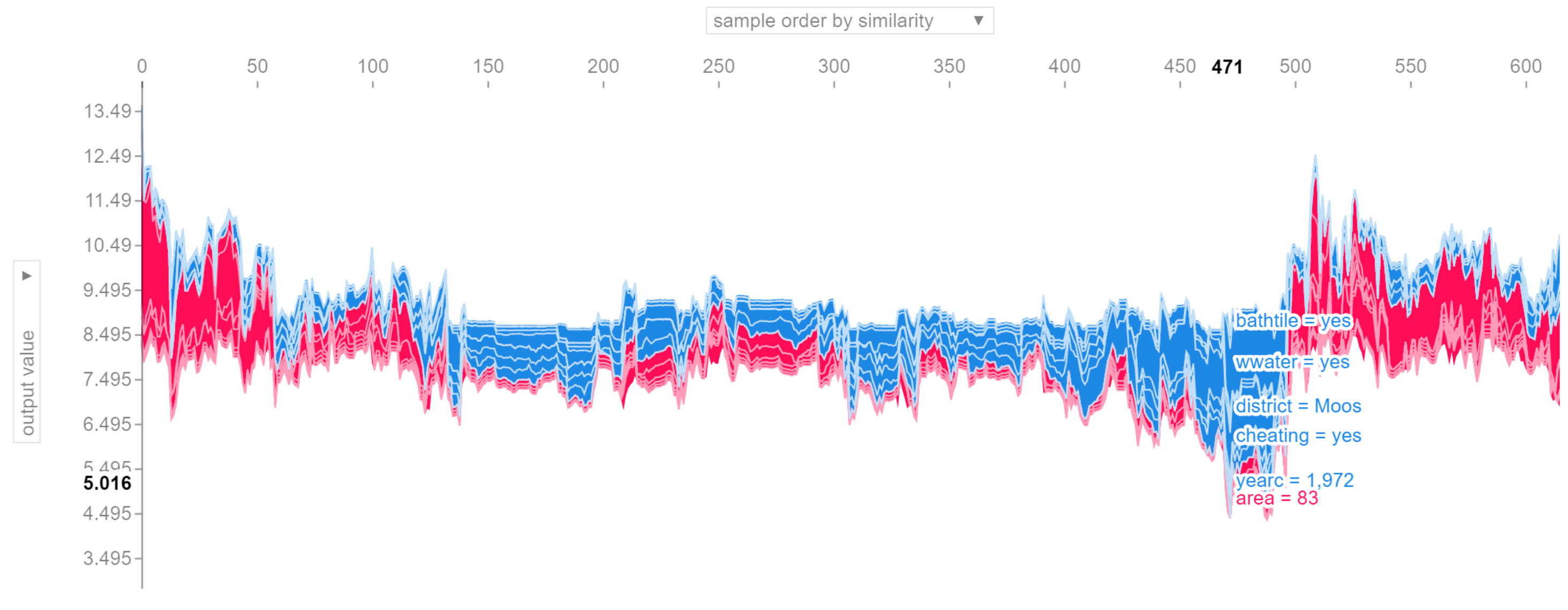}
	\caption{Shapley Values of $\mathbb{E}(Y|\mathbf{X} = \mathbf{x})$.}
	\label{fig:munich_shap_all_test}
\end{figure}

\noindent Besides the global attribute importance, the user might also be interested in local attribute importance for each single prediction individually. This allows to answer questions like 'How did the feature values of a single data point affect its prediction?' For illustration purposes, we select the first predicted rent of the test data set and present the local attribute importance for $\mathbb{E}(Y|\mathbf{X} = \mathbf{x})$ .

\begin{figure}[h!]
	\centering
	\includegraphics[width=0.8\linewidth]{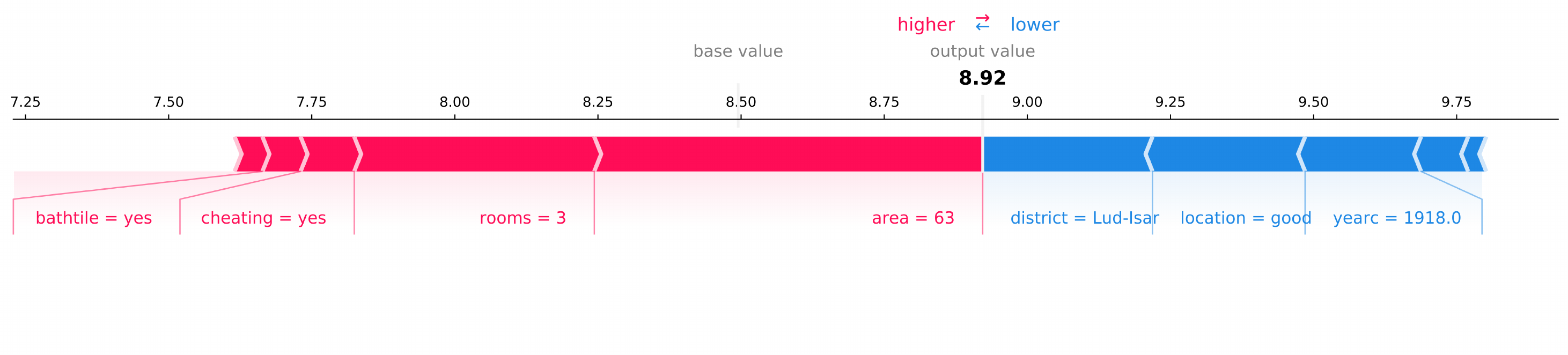}
	\caption{Local Shapley Value of $\mathbb{E}(Y|\mathbf{X} = \mathbf{x})$.}
	\label{fig:munich_shap_local}
\end{figure}

\begin{figure}[h!]
	\centering
	\includegraphics[width=0.8\linewidth]{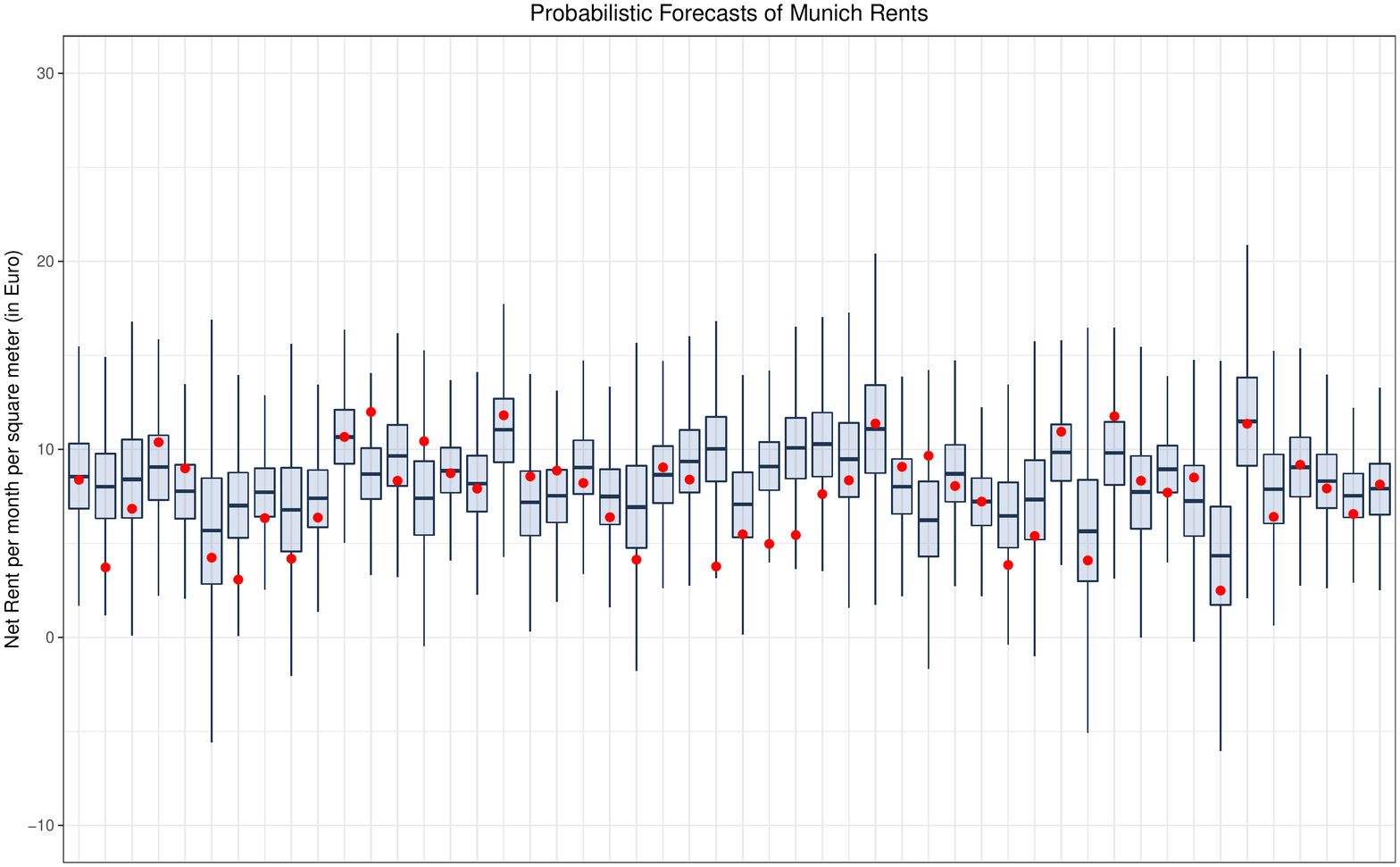}
	\caption{Boxplots of Probabilistic Forecasts of Munich Rents.}
	\label{fig:munich_forecast}
\end{figure}

\noindent As we have modelled all parameters of the Normal distribution, \texttt{CatBoostLSS} provides a probabilistic forecast, from which any quantity of interest can be derived. Figure \ref{fig:munich_forecast} shows a random subset of 50 predictions only for ease of readability. The red dots show the actual out of sample rents, while the boxplots visualise the predicted distributions.

\begin{figure}[h!]
	\centering
	\includegraphics[width=0.8\linewidth]{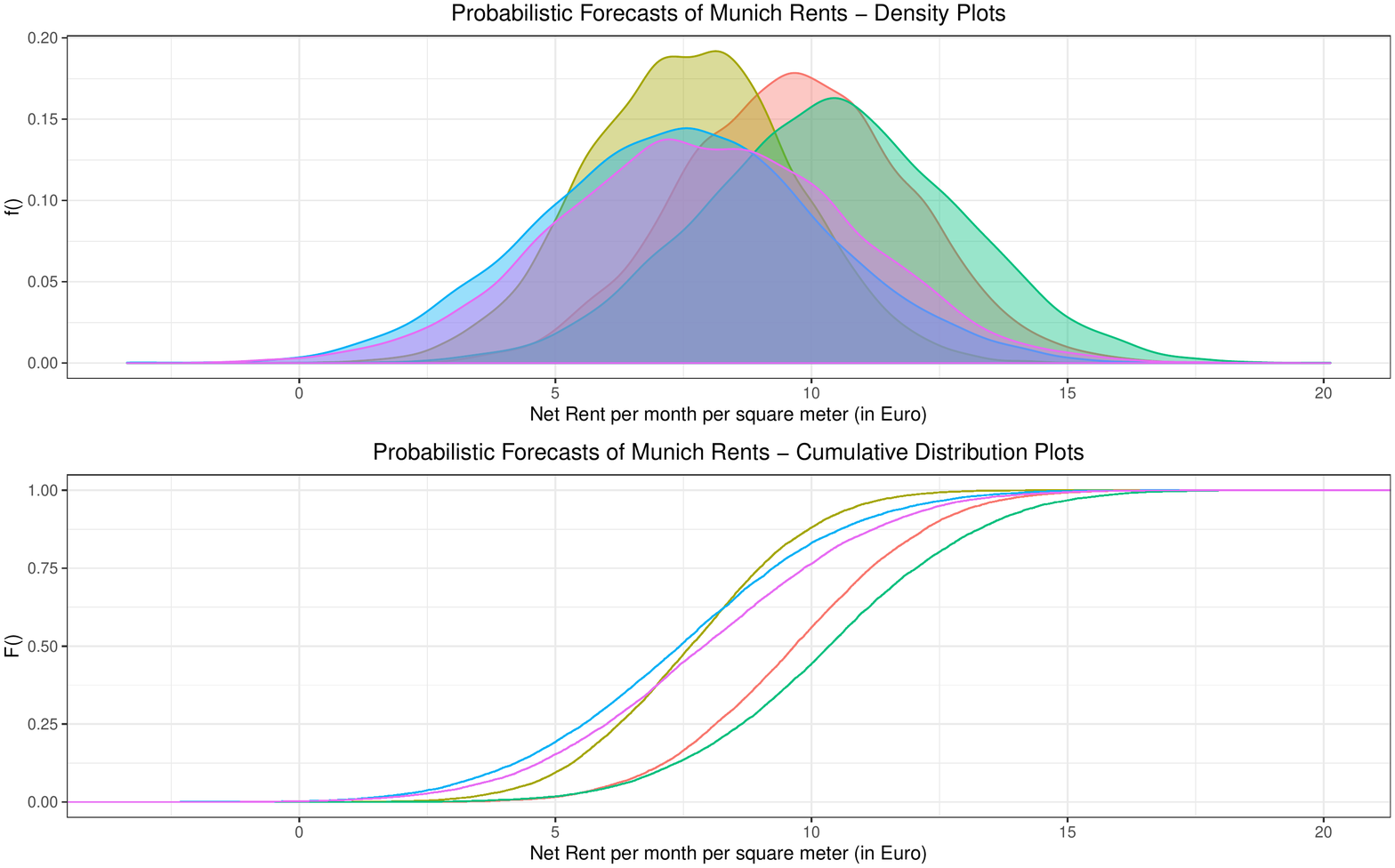}
	\caption{Density and Cumulative Distribution Plots of Probabilistic Forecasts of Munich Rents.}
	\label{fig:munich_dens}
\end{figure}

\noindent We can also plot a subset of the forecasted densities and cumulative distributions as shown in Figure \ref{fig:munich_dens}.

\subsubsection{Comparison to other approaches} \label{sec:comp}

To evaluate the prediction accuracy of \texttt{CatBoostLSS}, we compare the forecasts of the Munich rent example to the implementations available in $\it{XGBoostLSS}$, $\it{gamlss}$, $\it{gamboostLSS}$, $\it{blackboostLSS}$\footnote{$\it{blackboostLSS}$ is a gradient boosting approach using conditional inference trees as base-learners.}, as well as to the Bayesian formulation of GAMLSS implemented in $\it{bamlss}$ by \citet{Umlauf.2017}, to Distributional Regression Forests of \citet{Schlosser.2018, Schlosser.2019} implemented in $\it{distforest}$ and to $\it{ngboost}$ as introduced by \citet{Duan.2019}. For all approaches that can handle categorical information directly, we use factor coding. We evaluate distributional forecasts in Table \ref{tab:comp} using the average Continuous Ranked Probability Scoring Rules (CRPS) and the average Logarithmic Score (LOG), where lower scores indicate a better forecast, along with additional error measures evaluating the mean-prediction accuracy of the models.\footnote{Scoring rules are functions $S(\hat{F},\veclatin{y})$ that assess the quality of forecasts by assigning a value to the event that observations from a hold-out sample $\veclatin{y}$ are observed under the predictive distribution $\hat{F}$, with estimated parameter vectors $\hat{\veclatin{\theta}} = (\hat{\theta}_{1},\ldots,\hat{\theta}_{K})^{\prime}$. See \citet{Gneiting.2007} for details.}

\begin{table}[h!]
	\begin{center}
		\scalebox{0.8}{
			\begin{threeparttable}
				\caption{Forecast Comparison}
				\begin{tabular}{rrrrrrrrr}
					\toprule
					Metric     & CatBoostLSS & XGBoostLSS & gamboostLSS & GAMLSS & BAMLSS & DistForest & blackboostLSS & NGBoost \\  
					\midrule
					CRPS-SCORE & 1.1562 & \textbf{1.1415} & 1.1541 & 1.1527 & 1.1509 & 1.1554 & 1.2315 & 1.1634 \\   
					LOG-SCORE  & 2.1635 & \textbf{2.1350} & 2.1920 & 2.1848 & 2.1656 & 2.1429 & 2.7904 & 2.2226 \\ 
					MAPE       & 0.2492 & \textbf{0.2450} & 0.2485 & 0.2478 & 0.2478 & 0.2532 & 0.2650 & 0.2506 \\ 
					MSE 	   & 4.0916 & \textbf{4.0687} & 4.1596 & 4.1636 & 4.1650 & 4.2570 & 4.5977 & 4.2136 \\ 
					RMSE 	   & 2.0228 & \textbf{2.0171} & 2.0395 & 2.0405 & 2.0408 & 2.0633 & 2.1442 & 2.0527 \\ 
					MAE 	   & 1.6129 & \textbf{1.6091} & 1.6276 & 1.6251 & 1.6258 & 1.6482 & 1.7148 & 1.6344 \\ 
					MEDIAN-AE  & 1.3740 & 1.4044          & 1.3636 & \textbf{1.3537} & 1.3542 & 1.3611 & 1.4737 & 1.3574 \\ 
					RAE 	   & 0.7827 & \textbf{0.7808} & 0.7898 & 0.7886 & 0.7890 & 0.7998 & 0.8322 & 0.7932 \\ 
					RMSPE      & 0.3955 & \textbf{0.3797} & 0.3900 & 0.3889 & 0.3889 & 0.3991 & 0.4230 & 0.3950 \\ 
					RMSLE      & 0.2487 & \textbf{0.2451} & 0.2492 & 0.2490 & 0.2490 & 0.2516 & 0.2611 & 0.2507 \\ 
					RRSE       & 0.7784 & \textbf{0.7762} & 0.7848 & 0.7852 & 0.7853 & 0.7939 & 0.8251 & 0.7899 \\ 
					R$^{2}$    & 0.3942 & \textbf{0.3975} & 0.3841 & 0.3835 & 0.3833 & 0.3697 & 0.3192 & 0.3761 \\ 
					\bottomrule
				\end{tabular}
				\begin{tablenotes}
					\tiny
					\item \noindent Average Continuous Ranked Probability Scoring Rules (CRPS); Average Logarithmic Score (LOG); Mean Absolute Percentage Error (MAPE); Mean Square Error (MSE); Root Mean Square Error (RMSE); Mean Absolute Error (MAE); Median Absolute Error (MEDIAN-AE); Relative Absolute Error (RAE); Root Mean Square Percentage Error (RMSPE); Root Mean Squared Logarithmic Error (RMSLE); Root Relative Squared Error (RRSE); R-Squared/Coefficient of Determination (R$^{2}$). Best out-of-sample results are marked in bold (lower is better, except $R^{2}$).
				\end{tablenotes}
				\label{tab:comp}
		\end{threeparttable}}
	\end{center}
\end{table}

\noindent Comparing the results to its hyper-parameter tuned competitors, all measures show that \texttt{CatBoostLSS} provides a competitive forecast using default hyper-parameter settings.\footnote{We haven't performed any parameter tuning for Distributional Regression Forests in our comparison, as the runtime for a forest with $T$ = 1,000 trees took around 3.5 hours on a Windows machine. $\it{gamboostLSS}$ and $\it{blackboostLSS}$ are trained using parallelized 10-fold cross-validation to select the optimal number of iterations and $\it{XGBoostLSS}$ uses Bayesian Optimization for parameter tuning. $\it{NGBoost}$ was fit using a manual tuning of parameters, as to the knowledge of the author, currently no automated tuning is available. As such, the results of $\it{NGBoost}$ are prone to over-fitting, as the test data has been used to select the hyper-parameters. Tables \ref{tab:quant_loss} and \ref{tab:quant_loss_rank} further compare the different approaches across several quantiles using a quantile loss measure.} 

\subsubsection{Expectile Regression}

While GAMLSS require to specify a parametric distribution for the response, it may also be useful to completely drop this assumption and to use models that allow to describe parts of the distribution other than the mean. This may in particular be the case in situations where interest does not lie with identifying covariate effects on specific parameter of the response distribution, but rather on the relation of extreme observations on covariates in the tails of the distribution. This is feasible using Quantile and Expectile Regression. As with mean regression models, where the conditional mean is modelled as a function of covariates, both Quantile and Expectile Regression relate any specific quantile/expectile $\tau$ of the response to a set of covariates. Consequently, any desired point of the response distribution can be modelled, so that a dense grid of regressions yields a detailed description of the conditional distribution. Therefore, estimating and comparing parameter estimates across a different set of quantiles/expectiles allows for fully characterising the response distribution and for investigating the differential effect that covariates may have on different points of the conditional distribution. For our Munich rent analysis, Quantile/Expectile Regression yields additional insight compared to mean regression models, as they provide a richer description of the relationship between the rent of a flat and its attributing values for different values of $\tau$. 

\begin{figure}[h!]
	\centering
	\includegraphics[width=0.7\linewidth]{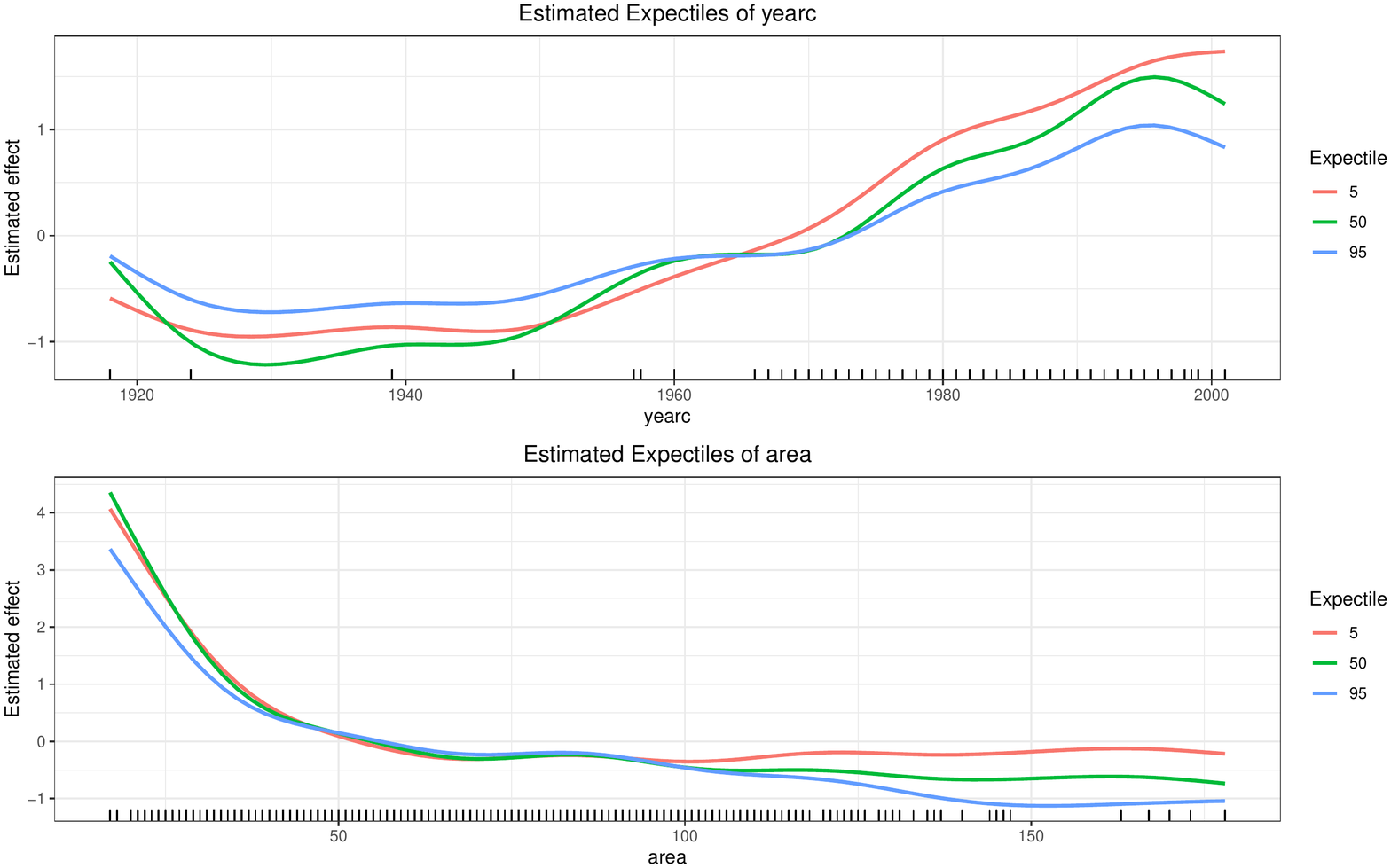}
	\caption{Estimated Partial Effects across different Expectiles.}
	\label{fig:expectile_pdp}
\end{figure}

As \texttt{CatBoostLSS} requires both the Gradient and Hessian to be non-zero, we illustrate the ability of \texttt{CatBoostLSS} to model and provide inference for different parts of the response distribution using Expectile Regression.\footnote{See \citet{Sobotka.2012} and \citet{Waltrup.2015} for further details on Expectile Regression.} Plotting the effects across different expectiles allows the estimated effects, as well as their strengths, to vary across the response distribution.\footnote{Even though excluded in theory, expectile crossing as shown in Figure \ref{fig:expectile_pdp} can occur, in particular with small data sets, as all expectiles are estimated separately. For suggestion on how to adjust the estimation process, we refer to \citet{Waltrup.2015} and the references therein.}

\noindent Investigation of the feature importances across different Expectiles allows to infer the most important covariates for each point of the response distribution so that, e.g., effects that are more important for expensive rents can be compared to those from affordable rents.

\begin{figure}[h!]
	\centering
	\includegraphics[width=0.7\linewidth]{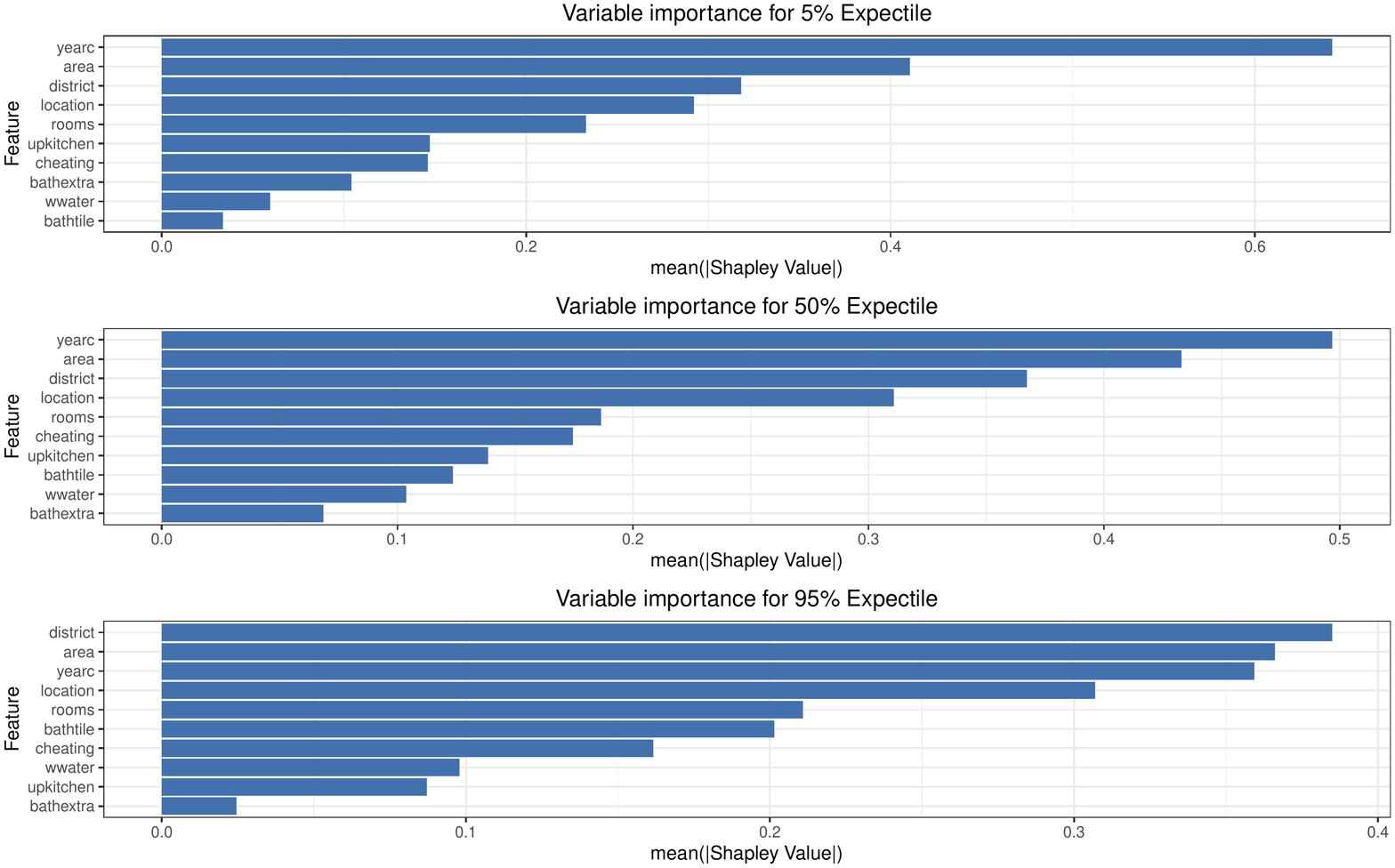}
	\caption{Mean Absolute Shapley Value across different Expectiles.}
	\label{fig:expectile}
\end{figure}

\section{Software implementation} \label{sec:implementation}

In its current implementation, \texttt{CatBoostLSS} is available in \texttt{Python} and will be made publicly available on the project Git-repo following this link \faGithub\href{https://github.com/StatMixedML/CatBoostLSS}{StatMixedML/CatBoostLSS}.

\section{Conclusion} \label{sec:conclusion}  

\begin{quote} 
	\it{There is indeed more to life than mean and variance. A good point at which to start is by replacing them by location and scale and noting that one reason for the stress on mean and variance is the implicit assumption of Gaussianity. Once the assumption of Gaussianity is dropped, \textbf{attention shifts to estimating [all of] the parameter in a distribution}.}\footnote{Emphasize added.} \citep{Harvey.2013}
\end{quote}

The language of statistics is of probabilistic nature. Any model that falls short of providing quantification of the uncertainty attached to its outcome is likely to yield an incomplete and potentially misleading picture. However, quantification of uncertainty in general and probabilistic forecasting in particular doesn't just provide an average point forecast, but it rather equips the user with a range of outcomes and the probability of each of those occurring. In an effort of bringing both disciplines closer together, this paper extends CatBoost to a full probabilistic forecasting framework termed \texttt{CatBoostLSS}. By exploiting its Newton boosting nature and the close connection between empirical risk minimization and Maximum Likelihood estimation, our approach models and predicts the entire conditional distribution from which prediction intervals and quantiles of interest can be derived. As such, \texttt{CatBoostLSS} provides a comprehensive description of the response distribution, given a set of covariates. By means of a simulation study and real world examples, we have shown that models designed mainly for prediction can also be used to describe and explain the underlying data generating process of the response of interest. 

\bibliography{references}  

\newpage

\setcounter{table}{0}
\renewcommand{\thetable}{A\arabic{table}}

\begin{appendices}

\section{Munich Rent Quantile Loss Comparison}

\begin{table}[h!]
	\begin{center}
		\scalebox{0.9}{
			\begin{threeparttable}
				\caption{Munich Rent Quantile Loss Comparison}
				\begin{tabular}{rrrrrrrrr}
					\toprule
					Quantile     & CatBoostLSS & XGBoostLSS & gamboostLSS & GAMLSS & BAMLSS & DistForest & blackboostLSS & NGBoost \\ 
					\midrule
					1 & 0.0145 & 0.0145 & 0.0167 & 0.0160 & 0.0157 & \textbf{0.0137}  & 0.0234 & 0.0177 \\ 
					5 & 0.0543 & 0.0522 & 0.0552 & 0.0544 & 0.0535 & \textbf{0.0502}  & 0.0630 & 0.0567 \\ 
					10 & 0.0904 & 0.0862 & 0.0913 & 0.0906 & 0.0900 & \textbf{0.0853} & 0.1004 & 0.0918 \\ 
					20 & 0.1390 & \textbf{0.1357}  & 0.1394 & 0.1392 & 0.1387 & 0.1377 & 0.1520 & 0.1406 \\ 
					30 & 0.1683 & \textbf{0.1668} & 0.1711 & 0.1712 & 0.1706 & 0.1707 & 0.1859 & 0.1735 \\ 
					40 & 0.1869 & \textbf{0.1867} & 0.1919 & 0.1915 & 0.1912 & 0.1921 & 0.2049 & 0.1928 \\ 
					50 & 0.1966 & \textbf{0.1961} & 0.1984 & 0.1981 & 0.1981 & 0.2009 & 0.2090 & 0.1992 \\ 
					60 & 0.1926 & 0.1913 & 0.1913 & \textbf{0.1911} & 0.1912 & 0.1950 & 0.2004 & 0.1921 \\ 
					70 & 0.1751 & 0.1737 & \textbf{0.1714} & 0.1719 & 0.1721 & 0.1741 & 0.1796 & 0.1729 \\ 
					80 & 0.1433 & 0.1410 & \textbf{0.1391} & \textbf{0.1391} & 0.1394 & 0.1403 & 0.1447 & 0.1402 \\ 
					90 & 0.0932 & 0.0904 & 0.0887 & 0.0886 & \textbf{0.0882} & 0.0889 & 0.0932 & 0.0892 \\ 
					95 & 0.0565 & 0.0540 & 0.0530 & 0.0528 & \textbf{0.0522} & 0.0541 & 0.0590 & 0.0544 \\ 
					99 & 0.0153 & \textbf{0.0145} & 0.0149 & 0.0152 & 0.0149 & 0.0167 & 0.0212 & 0.0163 \\ 
					\bottomrule
				\end{tabular}
				\begin{tablenotes}
					\tiny
					\item \noindent For a given quantile $\tau \in (0, 1)$, a target value $y$ and the $\tau$-quantile prediction $\hat{y}(\tau)$, the $\tau$-quantile loss is defined as $\mbox{QL}_{\tau}[y,\hat{y}(\tau)] = 2[\tau(y - \hat{y}(\tau)) \mathds{1}_{y - \hat{y}(\tau) > 0} + (1-\tau)(\hat{y}(\tau) - y)\mathds{1}_{y - \hat{y}(\tau)\leq 0}]$. We use a normalized sum of quantile losses $\sum^{n}_{i} \mbox{QL}_{\tau}[y_{i},\hat{y}_{i}(\tau)] / \sum^{n}_{i}|y_{i}|$. Best out-of-sample results are marked in bold (lower is better).
				\end{tablenotes}
				\label{tab:quant_loss}
			\end{threeparttable}
		}
	\end{center}
\end{table}

\begin{table}[h!]
	\begin{center}
		\scalebox{0.8}{
			\begin{threeparttable}
				\caption{Munich Rent Quantile Loss Comparison - Average Rank}
				\begin{tabular}{rrrrrrrrr}
					\toprule 
					& CatBoostLSS & XGBoostLSS & gamboostLSS & GAMLSS & BAMLSS & DistForest & blackboostLSS & NGBoost \\ 
					\midrule
					Average-Rank &  4.5385 & 2.8462 & 4.0000 & 3.5385 & \textbf{2.7692} & 4.3077 & 8.0000 & 6.0000 \\ 
					\bottomrule
				\end{tabular}
				\begin{tablenotes}
					\tiny
					\item \noindent Average rank calculation is based on Table \ref{tab:quant_loss}. Best out-of-sample results are marked in bold (lower is better).
				\end{tablenotes}
				\label{tab:quant_loss_rank}
			\end{threeparttable}
		}
	\end{center}
\end{table}

\end{appendices}

\end{document}